\documentclass[11pt]{article}

\usepackage[final]{acl}

\usepackage{times}
\usepackage{latexsym}

\usepackage[T1]{fontenc}

\usepackage[utf8]{inputenc}

\usepackage{microtype}

\usepackage{inconsolata}
\usepackage{amsmath}
\usepackage{amssymb}
\usepackage{enumitem}
\usepackage{xcolor}
\usepackage{graphicx}
\usepackage{cleveref}
\usepackage{subcaption}
\usepackage{booktabs}
\usepackage{multirow}
\usepackage{multicol}
\usepackage[table]{xcolor}
\usepackage{makecell}
\usepackage{ai-usage-card}

\usepackage{fvextra}    
\DefineVerbatimEnvironment{codeblock}{Verbatim}{
    fontsize=\small,
    baselinestretch=1.05,
    commandchars=\\\{\}, 
}
\definecolor{eebcb5}{HTML}{eebcb5}
\newcommand{\err}[1]{\colorbox{eebcb5}{#1}}
\definecolor{d87e7b}{HTML}{d87e7b}

\aiProjectName{Who Watches the Watchmen?}
\aiDomain{Paper}
\aiKeyApplication{Coding}

\aiModels{ChatGPT-5.2, GPT-5.2-Codex}
\aiGeneratingText{ChatGPT-5.2}
\aiImprovingContent{ChatGPT-5.2}
\aiPerspectiveWork{ChatGPT-5.2}
\aiGeneratingCode{GPT-5.2-Codex}
\aiRefactoringCode{GPT-5.2-Codex}
\aiComparingCode{GPT-5.2-Codex}
\aiWhyUse{Efficiency and Speed}
\aiMitigateErrors{We manually verified all outputs}
\aiMinimizeHarm{We carefully reviewed generated content}

\title{Who Watches the Watchmen?\\ Humans Disagree With Translation Metrics on Unseen Domains}

\author{
\textbf{Finn Schmidt\quad Jan Philip Wahle\quad
Terry Ruas \quad Bela Gipp}\\
University of Göttingen, Germany \\ 
\texttt{finn.schmidt@uni-goettingen.de}
}

\begin{document}
\maketitle

\begin{abstract}

Automatic evaluation metrics are central to the development of machine translation systems, yet their robustness under domain shift remains unclear. Most metrics are developed on the Workshop on Machine Translation (WMT) benchmarks, raising concerns about their robustness to unseen domains. Prior studies that analyze unseen domains vary translation systems, annotators, or evaluation conditions, confounding domain effects with human annotation noise.
To address these biases, we introduce a systematic multi-annotator \textbf{C}ross-\textbf{D}omain \textbf{E}rror-\textbf{S}pan-\textbf{A}nnotation dataset (CD-ESA), comprising 18.8k human error span annotations across three language pairs, where we fix annotators within each language pair and evaluate translations of the same six translation systems across one seen news domain and two unseen technical domains. 
Using this dataset, we first find that automatic metrics appear surprisingly robust to domain-shifts at the segment level (up to 0.69 agreement), but this robustness largely disappears once we account for human label variation.
Averaging annotations increases inter-annotator agreement by up to +0.11.
Metrics struggle on the unseen chemical domain compared to humans (inter-annotator agreement of 0.78–0.83 vs. 0.96). 
We recommend comparing metric–human agreement against inter-annotator agreement, rather than comparing raw metric–human agreement alone, when evaluating across different domains.
\end{abstract}

\section{Introduction}

Machine translation aims to break barriers between different cultures, supporting humanity’s ability to collaborate and share knowledge \citep{Steigerwald2022OvercomingLBA}. 
To quantify which translation methods are ``good'', we need robust quality estimation (QE) metrics that can assess translation quality based solely on the source and candidate translation, as reference translations are often unavailable when exploring new domains \citep{mathur-tangled}. 
Those QE metrics enable large-scale experimentation and frequent model comparisons that would be too costly with human annotation \citep{Chaganty2018ThePOA}.
  \, Although many studies rely on evaluation metrics as the \textit{watchmen} for progress in the field, there are valid concerns whether these metrics accurately quantify progress as intended, especially in low-resource settings \citep{kocmi2021ship}.


\begin{figure}
    \centering
    \includegraphics[width=\columnwidth]{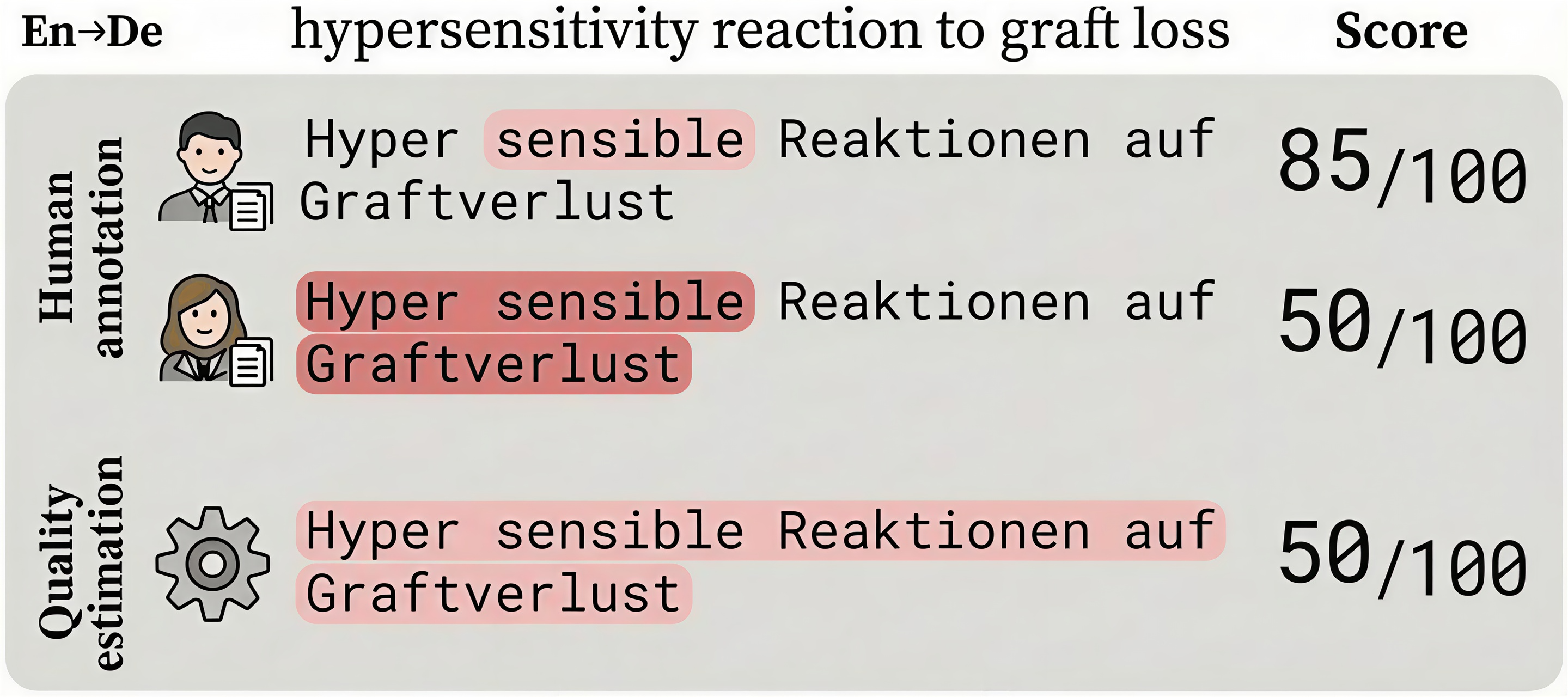}
    \caption{Metrics may show unexpected behavior on unseen domains, reflecting domain familiarity over translation quality, while variability in human judgments (e.g., overlooking errors) can mask these effects.}
    \label{fig:placeholder}
    \vspace*{-4mm}
\end{figure}

Most metric development and validation is centralized around the WMT shared tasks \citep{lavie2025findings}. 
 The WMT shared tasks provide a large, annually refreshed human-evaluated benchmark of bilingual text drawn from multiple high-resource domains such as news, literary, social, and spoken sources. 
This has raised concerns that metric performance degrades under conditions outside this distribution, for example on medical data. 
Few studies have directly examined QE metric robustness on unseen domains by collecting new human annotations tailored to out-of-domain evaluation \citep{zhang-etal-2025-good, zouhar}. 
However, comparing domains between studies is difficult and assessments are often done by different human evaluators, raising two methodological gaps to motivate this study. 
First, human judgments are known to be inconsistent \citep{belz2023missing}, especially at the segment level \citep{plank2022problem}. 
Individual ratings contain noise that can act as a bottleneck for reliable metric evaluation (illustrated in \Cref{fig:placeholder}). 
Second, studies that change annotator pools, translation methods, and annotation conditions across datasets make it challenging to compare results and attribute performance differences solely to domain shift \citep{zouhar}. 
Therefore, raw agreement or correlation scores are difficult to interpret, as they confound metric behavior with domain-specific annotation difficulty \citep{agrawal2024automaticmetricsassesshighquality}.
We argue that variation in human label consistency across domains and annotator pools makes machine translation scores difficult to compare. In effect, we need better ways for assessing the QE metrics in out-of-domain settings, this \textit{watching the watchmen}.

In this study, we introduce the \textbf{C}ross-\textbf{D}omain \textbf{E}rror-\textbf{S}pan-\textbf{A}nnotation dataset (CD-ESA) dataset with 18.8k human error span annotations from 2 or 4  annotators per language pair on English-Chinese (en-zh), English–German (en-de), and English-Korean (en-ko) translations across three domains: WMT23 news, public medical documents (Emea), and proprietary pharmaceutical and chemical data (PharmaChem).
We ensure fair cross-domain comparison by evaluating translations of the same MT systems and relying on the same annotators per language pair across all domains. 
Each segment is annotated by multiple human annotators for consistent measurement of inter-annotator agreement and fair comparison of annotation difficulty across domains. 
Since human labels can vary substantially \citep{plank2022problem, Sarti_2025}, we examine how averaging  human annotations reduces this variability. 
We also propose a grouping strategy that combines $n$ segments, reducing annotation costs and inconsistency by requiring fewer human judgments. 
To avoid data contamination, particularly for LLM-as-a-judge approaches \citep{kocmi-federmann-2023-large}, we derive our unseen-domain dataset from proprietary company data of shift worker log entires with technical terms and specific jargon.

We first find that QE metrics appear surprisingly robust at the segment level on unseen domains (up to 0.69 agreement), often within 0.04 of inter-annotator agreement, but this robustness largely disappears once we account for human label inconsistency. Averaging annotations increases inter-annotator agreement by up to +0.11. Metrics struggle in the unseen PharmaChem domain, inter-annotator agreement reaches 0.96, while metrics often achieve only 0.78–0.87. Metrics underestimate translation quality by up to 0.37 normalized score points and sometimes overprefer open-source MT systems. These effects are strongest for fine-tuned metrics such as \textsc{xComet\_QE} \cite{guerreiro2024xcomet}, while LLM-as-a-judge approaches show greater robustness.

Our CD-ESA annotations for WMT and Emea across all three langauge pairs are publicly available.\footnote{https://huggingface.co/datasets/FinnSchmidt/CD-ESA} 
PharmaChem contains proprietary data and cannot be released.

\medskip
\noindent\textbf{Key Contributions:}
\begin{itemize}[leftmargin=*,label={\color{teal}$\blacktriangleright$},itemsep=0.15em]
\item \textbf{A Cross-Domain ESA Dataset.}
We introduce \textbf{CD-ESA}, a multi-annotator \textbf{C}ross-\textbf{D}omain \textbf{E}rror \textbf{S}pan \textbf{A}nnotation dataset for three (unseen) domains under consistent evaluation conditions. 




\item \textbf{Disentangling metric robustness from human label inconsistency.}
To reduce human label inconsistency, we propose and evaluate a cost-effective alternative that groups $n$ segments into a single evaluation unit.

\item \textbf{Analysis of metric weaknesses under domain shift.}
We analyze the reasons why metrics fail under domain shift.

\end{itemize}
\section{Related Work}

\paragraph{Quality estimation} (QE) predicts translation quality without reference translations \citep{Specia2009EstimatingTS}, making it key for low-resource settings, where references are often not available. Most QE research follows two paradigms. First, methods fine-tune encoder-based models on human judgments, including CometKiwi \citep{rei2023scaling}, MetricX \citep{juraska-etal-2024-metricx}, and xComet \citep{guerreiro2024xcomet}. Remedy \citep{tan2025remedy} departs from regression-based training by learning relative quality from pairwise preferences, improving performance on standard benchmarks. Second, methods uses LLM-as-a-judge approaches, such as GEMBA-MQM \citep{kocmi2023gemba}, which assess translation quality via prompting \citep{kocmi-federmann-2023-large, fernandes2023devilerrorsleveraginglarge}.
While both paradigms report strong results on benchmark data, their robustness on new domains remains underexplored.

\paragraph{Meta-evaluation of QE}
Metric development and validation are strongly shaped by the WMT Metrics Shared Tasks, where submitted metrics are ranked by agreement with newly collected human judgments \citep{freitag2022results, freitag2023results, freitag-etal-2024-llms, lavie2025findings}. These benchmarks focus on a limited set of high-resource domains (e.g., news, literary text, reviews) and language pairs (e.g., en–de, en–zh). Over time, WMT has produced large human-annotated datasets using standardized evaluation protocols, such as Direct Assessment (DA) \citep{graham2013continuous}, Multidimensional Quality Metrics (MQM) \citep{freitag2021mqm}, and ESA \citep{kocmi2024error}. As a result, most state-of-the-art metrics train on WMT-derived judgments and achieve near-human agreement on WMT test sets \citep{proietti2025has}.
We argue that such results show adaptation to WMT-style distributions. Our work evaluates metrics under matched conditions but outside this distribution.


\paragraph{Metrics on unseen domains} Only few studies evaluate metrics using newly collected annotations on unseen domains. \citet{zouhar} report degraded metric performance out of domain, though updated analyses show sensitivity to label imbalance (\Cref{apd:acc_bio}). Further, they assume BLEU as a domain-invariant reference, despite evidence of domain-dependent performance (\Cref{apd:acc_bio}). Moreover, prior studies vary MT systems, annotators, or annotation protocols across domains, confounding domain effects with human variability. Recent work on QE in the literary domain \citep{zhang-etal-2025-good} focuses on human evaluation reliability rather than cross-domain metric robustness. 


Prior work highlights challenges in evaluating metrics on unseen domains but typically varies annotators, MT systems, or evaluation setups across datasets, making it difficult to isolate true domain effects from annotation noise. This lack of controlled cross-domain evaluation leaves it unclear whether observed performance drops reflect genuine metric weaknesses or differences in human labeling consistency.

\section{Methodology}

We create the CD-ESA dataset to evaluate MT metrics under domain shift by translating data from three domains with the same MT systems, and collecting multiple human error span annotations per segment.
We assess metric–human agreement at segment and system levels using the WMT25 \citep{lavie2025findings} protocols and compare it with inter-annotator agreement. 
This design allows us to disentangle metric robustness from human label inconsistency and to analyze systematic biases in metrics across unseen domains.

\subsection{Creation \& annotation of CD-ESA}
\label{subsec:dataset}



To create our CD-ESA dataset, we collect English source sentences from three domains:

\noindent \textbf{(i) WMT23 (in-domain)}
A randomly sampled subset of English news data from the metrics shared task of WMT23 consisting of well-edited journalistic text from a high-resource news domain on which most automatic metrics are developed and validated.

\noindent \textbf{(ii) PharmaChem (unseen, proprietary)}
A proprietary industrial dataset of English log entries written by shift workers\footnote{Shift workers operate industrial production processes and document processes, incidents, and handovers across shifts.} in pharmaceutical and chemical plants. The data contains domain-specific terminology, abbreviations, and short, non-fluent utterances. As proprietary data, PharmaChem is contamination-free for LLM-based evaluation.

\noindent \textbf{(iii) Emea (unseen, public proxy) }
A parallel corpus derived from documents from the European Medicines Agency \citep{emea}. The data contains medical and regulatory text with specific terminology and abbreviations. As a public proxy for PharmaChem, we retain only segments classified as similar to PharmaChem using a trained binary domain classifier. We trained a SciBERT domain classifier on an 80k balanced dataset (50\% unrelated domains, 50\% Medicine/Pharma and sampled PharmaChem data). We then applied it to multilingual corpora, identified Emea as the closest match, and retained sentences with classifier score > 0.5 as a public proxy (for more details see \Cref{apd:Domain_Data_filter}).



We translate all source sentences from English into German by the same six MT systems: GPT-4o \citep{openai2024gpt4o}, DeepL \citep{deepl}, Azure MS \citep{azuremt}, X-ALMA \citep{xu2024x}, Tower-Instruct-v2 \citep{rei2024tower}, and a Tower model we fine-tune on PharmaChem data. This results in a total of 1354 translations (\Cref{tab:esa-overview-avg}). 

To ensure comparability and avoid introducing inference-related artifacts, all open-source MT systems use beam decoding \citep{Sutskever2014SequenceTS}. 
This choice avoids metric-hacking effects of other decoding strategies, such as Minimum Bayesian Risk decoding \citep{freitag2022high, tomani2024quality, pombal2025adding}.

We annotate all translations using a total of five annotators and at least two annotators per segment, enabling estimates of inter-annotator agreement across domains.
All annotations follow the Error Span Annotation (ESA) protocol \citep{kocmi2024error} and are collected using Label Studio\footnote{\url{https://labelstud.io}}.

For en-de translations, annotators are native German speakers (four female, one male), aged between 20 and 25, and employed as working students with prior experience in linguistic annotation tasks. For en-ko and en-zh translations, annotators are professional translators with native proficiency in Korean and Chinese. They are recruited through a certified translation vendor.

\Cref{tab:esa-overview-avg} summarizes dataset statistics. PharmaChem is the most challenging domain, with 2.8 errors per segment on average, compared to 1.4 on WMT23 and 0.8 on Emea. The lower error rate on Emea is partly explained by shorter segments (22.4 vs. 38 tokens for WMT23 and PharmaChem).




Following human annotation, we evaluate three categories of automatic metrics:

\noindent\textbf{(i) Supervised fine-tuned (SFT) metrics}, including xCOMET-XL\_QE, MetricX-24-XL\_QE, and COMETKIWI-23-XL. Unless stated otherwise, reported SFT metrics results correspond to the mean performance across these three metrics.

\noindent\textbf{(ii) Remedy}, a preference-based metric that outperforms fine-tuning on recent WMT benchmarks.

\noindent\textbf{(iii) GEMBA}, an \textbf{LLM-as-a-judge approach} using the GEMBA-ESA-QE prompt\footnote{\url{https://github.com/MicrosoftTranslator/GEMBA}} with GPT-4o\footnote{We use the GPT-4o API version released on 2024-12-01.}.

Since reference translations are often unavailable in low-resource and domain-specific settings, we focus exclusively on the \textbf{QE capabilities of these metrics}. 
All evaluated metrics are explicitly trained or configured for reference-free evaluation.


\begin{table}[t]
  \centering
  \small
  \begin{tabular}{lccc}
    \toprule
    Metric & Emea & \makecell{Pharma-\\Chem} & WMT23 \\
    \midrule
    \# Annotated segments & \textbf{3060} & \textbf{3808} & \textbf{1668} \\
    \# Sources & 528 & 652 & 249 \\
    Avg.\ words / source & 28.6 & 36.7 & 46.5 \\
    \makecell{\# Annotators per segment \\ for en-de/en-ko/en-zh} & \textbf{4/2/2} & \textbf{2/2/2} & \textbf{4/2/2} \\
    Avg.\ ESA score & 80.9 & 65.1 & 83.8 \\
    Avg.\ errors / seg & 1.4 & 2.9 & 1.6 \\
    Minor errors [\%] & 58.8 & 42.1 & 73.0 \\
    Major errors [\%] & 41.2 & 57.9 & 27.0 \\
    \bottomrule
  \end{tabular}
  \caption{Dataset overview of our new dataset CD-ESA.}
  \label{tab:esa-overview-avg}
  \vspace*{-4mm}
\end{table}

\subsection{Meta-Evaluation}
\label{subsec:Meta}

To assess automatic metrics on the CD-ESA dataset, we follow the WMT25 Metrics shared task evaluation and measure metric-human agreement at system- and segment-level. 

\paragraph{Segment-level evaluation.}


For segment-level evaluation, we use pairwise accuracy with tie calibration ($\mathrm{acc}_{eq}$) \citep{deutsch2023ties}. 
Specifically, $\mathrm{acc}_{eq}$ counts how often a metric $m$ ranks pairs of translations of the same source segment in the same order as a human annotator $h$, while accounting for tied scores.


\paragraph{System-level evaluation.}
We evaluate system-level metric performance using Soft Pairwise Accuracy (SPA)~\cite{thompson2024improving}. 
SPA measures how well a metric agrees with humans on the relative ranking of MT systems while accounting for the confidence of these preferences. Computational details of $acc_{eq}$ and SPA are in \Cref{apd:meta-eval}.

\paragraph{Oracle bounds} We also establish performance bounds ensuring fair comparison across domains. 
We measure inter-annotator agreement using the same metrics ($\mathrm{acc}_{\mathrm{eq}}$, SPA) to define an upper bound. 
As lower bounds, we include a dummy metric that assigns random scores and a sentinel candidate metric \citep{perrella2024guardians} that has access only to the candidate translation and not the source. Thus, this metric is unable to evaluate the quality of machine-translated text properly.

Unless stated otherwise, we compute evaluation scores by treating each human annotator as the ground truth individually and averaging the resulting values. 
For inter-annotator agreement, we compute agreement over all possible annotator pairs and report the average.

\paragraph{Bias analysis and implementation.}
To analyze potential biases, we additionally examine scores assigned by metrics and humans across MT systems and domains directly. 
This allows us to detect preferences for specific systems and systematic score shifts across domains.
We conduct parallel analyses of human scoring to distinguish effects caused by human label inconsistency from metric biases.

\section{Experiments}

Our experiments investigate how well automatic MT evaluation metrics generalize to unseen domains and identify factors that limit their reliability under domain shift.

First, we assess whether current metrics exhibit degraded performance on unseen domains. 
We compare metric–human agreement at the segment and system levels across seen and unseen domains, following the WMT25 meta-evaluation protocol.
Interpreting metric–human agreement relies on human judgments as ground truth. 
However, human annotations are known to exhibit substantial label inconsistency \citep{plank2022problem}. This raises the question of whether human noise may conceal differences in metric behavior across domains.
Second, we investigate the role of human label inconsistency in cross-domain evaluation. 
By averaging human scores and grouping multiple segments, we reduce annotation noise and obtain a cleaner evaluation signal. 
This allows us to test whether domain-specific effects become more apparent once human noise is mitigated.
Third, we analyze the reasons metrics fail to align with human judgments on unseen domains. 
We focus on identifying systematic biases and domain-specific effects that explain mismatches between metric and human evaluations.

\subsection{QE metrics appear robust on unseen domains}
\label{subsec:SPA_acc_eq}


\begin{figure}[t]
  \centering

  \begin{subfigure}{\columnwidth}
    \centering
    \includegraphics[width=\linewidth]{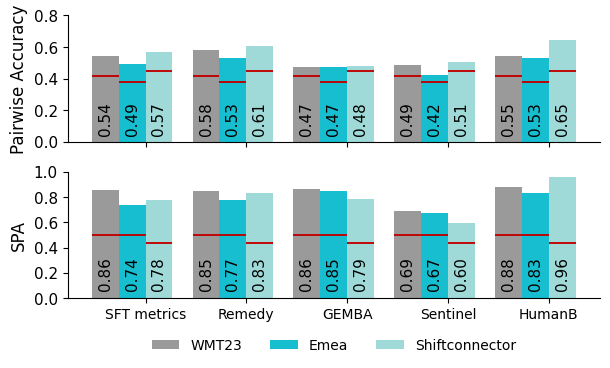}
    \caption{English-German}
  \end{subfigure}

  \begin{subfigure}{\columnwidth}
    \centering
    \includegraphics[width=\linewidth]{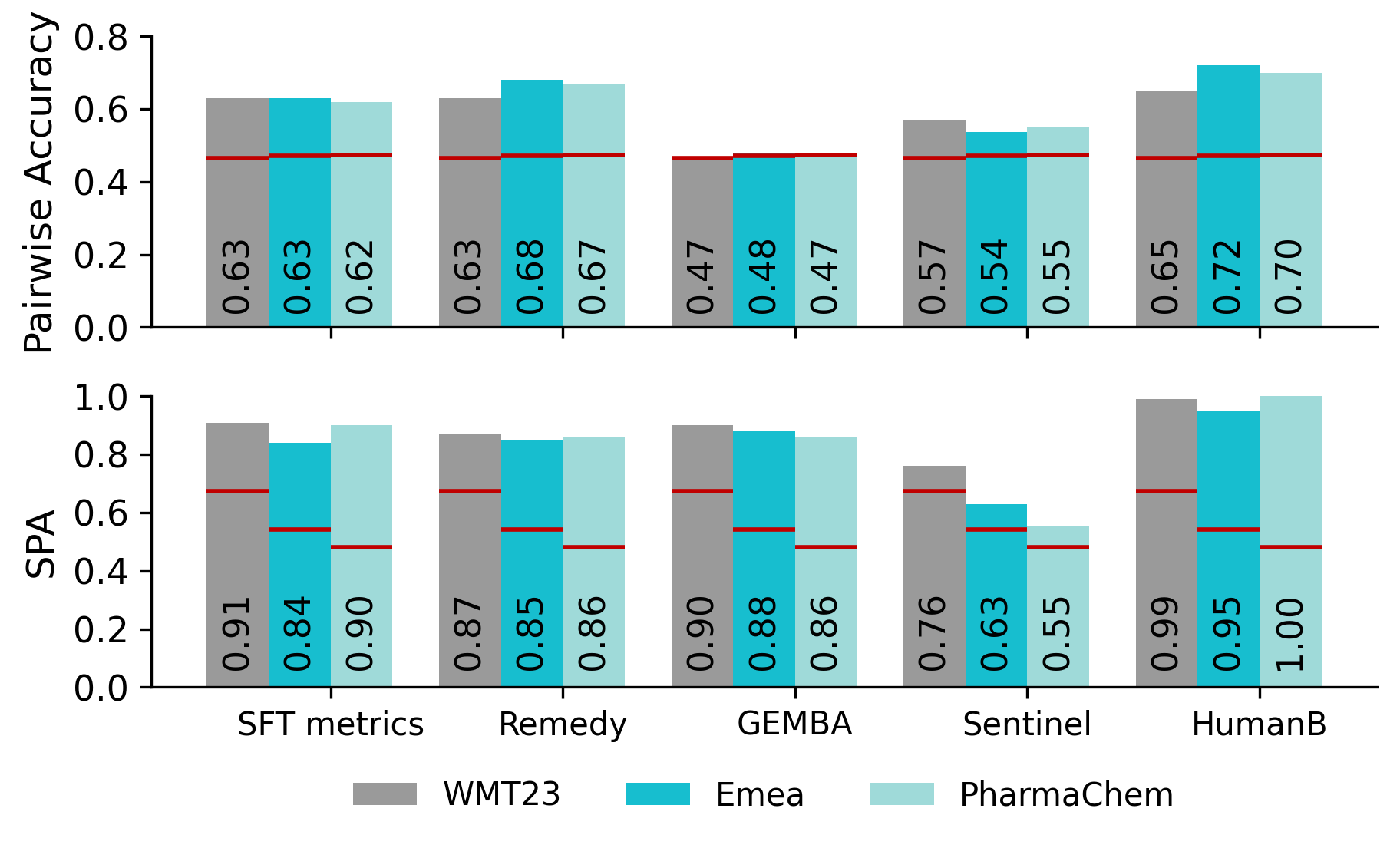}
    \caption{English-Korean}
  \end{subfigure}

  \begin{subfigure}{\columnwidth}
    \centering
    \includegraphics[width=\linewidth]{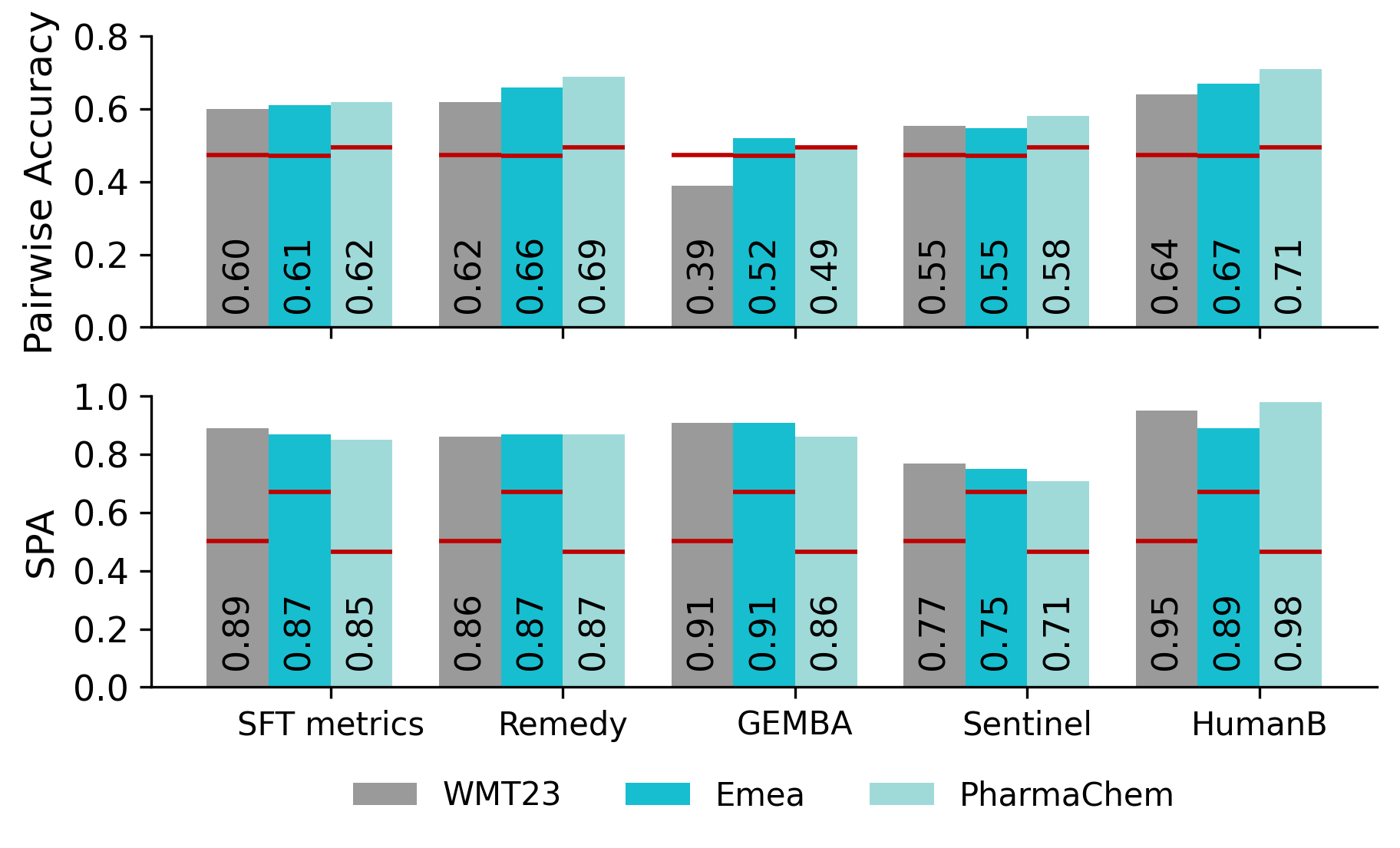}
    \caption{English-Chinese}
  \end{subfigure}

  \caption{(Soft)Pairwise accuracies of metric–human and human–human (humanB) agreement on CD-ESA. 
  The red line is the performance of a random baseline.}
  
  \vspace*{-4mm}
  \label{fig:acc_plain}
\end{figure}

To compare QE metric performance across seen and unseen domains, we measure segment- and system-level metric–human agreement on our CD-ESA dataset across three domains: WMT23, Emea, and PharmaChem (\Cref{subsec:dataset}) across the three language pairs en-de, en-ko, en-zh. \Cref{fig:acc_plain} reports segment-level $acc_{eq}$ and system-level SPA. 



At the segment level, metrics show no degradation on unseen domains. Across all three language pairs, SFT metrics and Remedy remain stable or even slightly improve on the unseen PharmaChem domain (0.60→0.62 and 0.62→0.69 on en-zh) compared to WMT23. GEMBA also shows consistent performance across domains (0.47-0.48 accuracy on en-de, en-ko for all domains). Overall, segment-level metric–human agreement remains stable across domains.

Comparison to inter-annotator agreement further supports this robustness. Across all domains and language pairs, at least one metric is within approximately 0.03–0.05 of human–human agreement. For example, Remedy closely tracks human agreement across settings (e.g., differences of $\leq$0.03 on WMT23, $\leq$0.04 on Emea, and $\leq$0.04 on PharmaChem). In some cases, metric–human agreement even exceeds inter-annotator agreement (e.g., Remedy: 0.58 vs.\ 0.55 on WMT23 en-de), suggesting that human label inconsistency may affect reliability \citep{plank2022problem}.



At the system level, a different pattern emerges. On WMT23, metrics achieve high agreement (typically $\sim$0.85–0.91), close to inter-annotator agreement across language pairs. On Emea, GEMBA remains competitive (e.g., $\sim$0.85–0.91) and in some cases matches or slightly exceeds human agreement, while SFT metrics and Remedy tend to lag behind (e.g., $\sim$0.74–0.87 depending on the language pair). On PharmaChem, inter-annotator agreement is consistently high (up to $\sim$0.96–1.00), but metrics fall behind across all language pairs, with SPA values typically between $\sim$0.78-0.87. This widening gap indicates that, once human label noise is reduced, metric weaknesses on unseen domains become more apparent at the system level despite their apparent robustness at the segment level.

\begin{figure}[t]
  \includegraphics[width=\columnwidth]{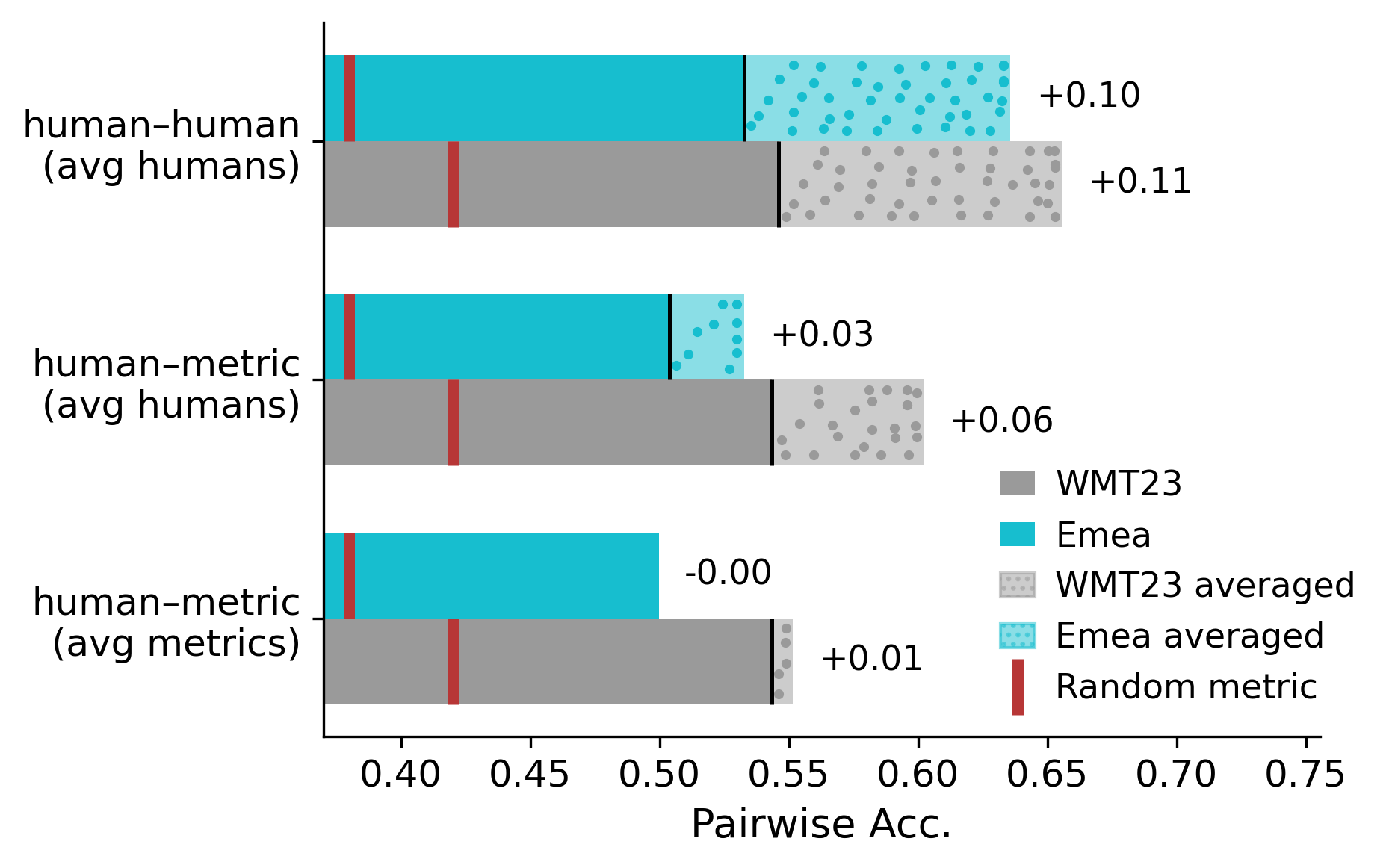}
  \vspace*{-5.5mm}
  \caption{Pairwise accuracy gains from score aggregation on WMT23 and Emea.
Bars report changes in $acc_{eq}$ when replacing single with averaged scores.
}
  \label{fig:averaging}
  \vspace*{-4mm}
\end{figure}

\subsection{Averaging human annotations reveals metric weaknesses on unseen domains} 
\label{sec:averaging}
Here, we want to understand whether the apparent cross-domain robustness observed in \Cref{subsec:SPA_acc_eq} reflects metric generalization or is instead driven by human label inconsistency. 
We evaluate the effect of averaging multiple human annotations as a noise-mitigation strategy and propose a second, cost-effective strategy that groups $n$ segments into a single evaluation unit. As we have four instead of two annotators per segment only for en-de, we experiment only with en-de data in this experiment. 

\paragraph{Aggregating multiple annotations.} We assume that each translation has an underlying ground-truth quality score and that annotators approximate this score with independent noise \citep{Graham2015AccurateEOA}. Under this assumption, averaging multiple annotations for the same segment should reduce noise. 
As a result, we expect both inter-annotator and metric–human agreement to increase as more annotations are aggregated \citep{Sarti_2025}.

To enable averaging across heterogeneous scoring scales, we first z-normalize all scores for each annotator and each metric separately, following prior work \citep{bojar-etal-2017-results}. 
We first examine inter-annotator agreement. 
Specifically, we compare $acc_{eq}$ at the segment level between two individual annotators (single–single) to agreement computed between two groups of annotators, where each group’s scores are averaged over two annotators (pair–pair). 
In both cases, we average the results over all possible single–single and pair–pair annotator combinations.
Averaging human annotations leads to a consistent improvement across domains, suggesting that human label inconsistency is largely domain-independent. 

As shown in the top bars of \Cref{fig:averaging}, averaging yields consistent gains across domains: inter-annotator agreement increases from 0.55 to 0.66 on WMT23 (+0.11) and from 0.53 to 0.63 on Emea (+0.10), indicating largely domain-independent human label noise.


Next, we analyze the average metric–human agreement by replacing a single human annotation with the average of four human annotations as ground truth (middle bars in \Cref{fig:averaging}). On WMT23, agreement increases from 0.54 to 0.60 (+0.06). On the unseen Emea domain, gains are smaller, from 0.50 to 0.53 (+0.03), despite comparable reductions in human noise. This divergence indicates that once annotation noise is reduced, metric weaknesses on unseen domains become more apparent.

Finally, we examine average metric–human agreement when averaging scores from four metrics instead of using a single metric score (bottom bars in \Cref{fig:averaging}). In contrast to human averaging, this yields no meaningful improvement. The absence of gains indicates that current metrics exhibit correlated errors rather than independent noise, reflecting shared biases.

\begin{figure}[t]
  \includegraphics[width=\columnwidth]{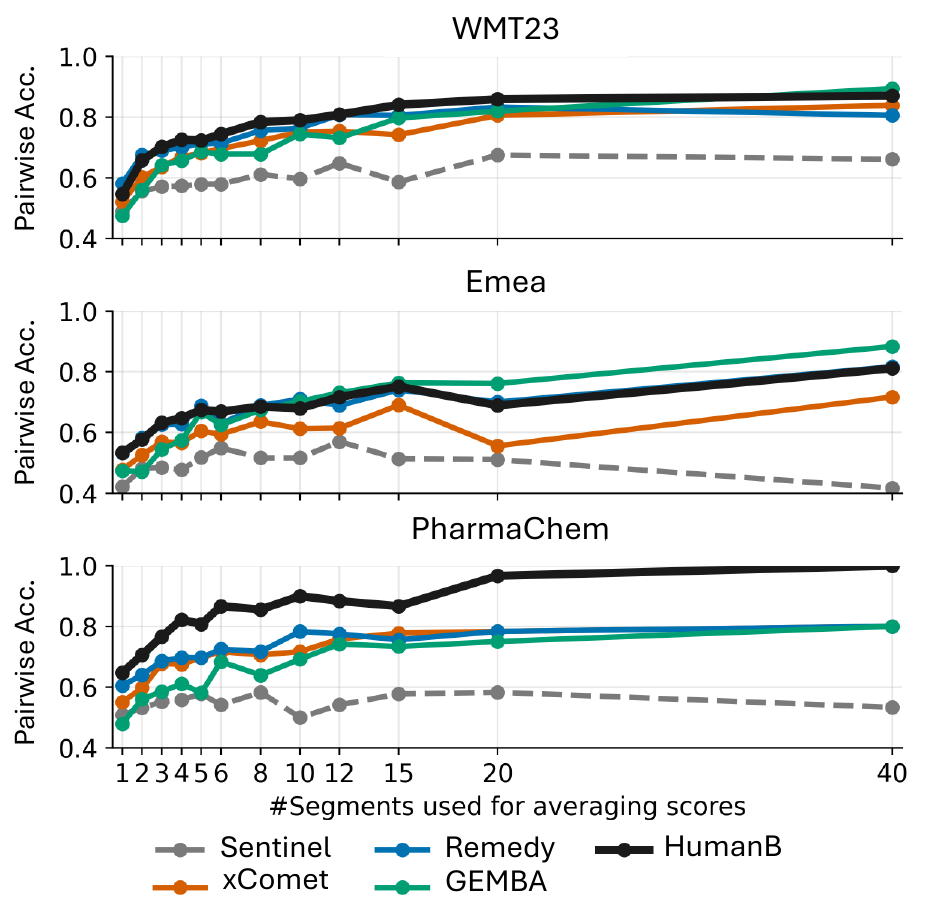}
  \caption{Pairwise accuracy for grouping segments. The x-axis shows number of segments (\(n\)), the y-axis shows pairwise accuracy. We plot inter-annotator agreement (HumanB) and metric–human agreement for several metrics with lower-bound (sentiel).}
  \label{fig:grouping}
  \vspace*{-4mm}
  \end{figure}

\begin{figure*}[t]
    \centering
    \includegraphics[width=0.9\textwidth]{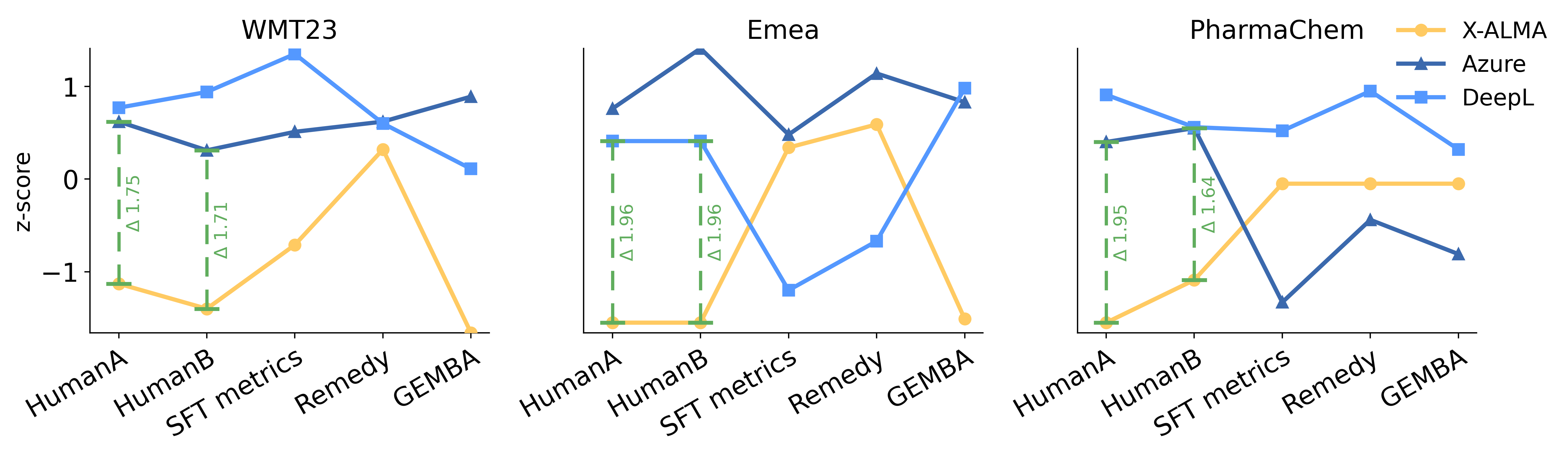}
    \caption{Visualization of the ranking of three MT systems. The x-axis lists the ranks and the y-axis shows the corresponding z-scores assigned by the ranker. Green dotted $\Delta$ markers in the HumanA/HumanB columns indicate the score gap between X-ALMA and the closed-source systems under human judgments.}
    \label{fig:human_metric_rankers}
    \vspace{-4mm}
\end{figure*}

\paragraph{Grouping $n$ segments.}
Since collecting multiple human annotations per segment is costly and often infeasible for large datasets, we propose an alternative noise-mitigation strategy: \textbf{grouping $n$ segments}. 
Instead of averaging multiple annotations for the same segment, we group $n$ segments translated by the same MT system and average their scores. 
This yields an intermediate evaluation setting between segment-level evaluation ($n=1$) and full system-level evaluation (large $n$).

\Cref{fig:grouping} shows the results.
For WMT23 with $n=1$, the inter-annotator agreement is relatively low with 0.54, reflecting high inconsistency in human segment-level labels. 
As $n$ increases, inter-annotator accuracy rises sharply. 
Grouping only three segments already increases human–human pairwise accuracy from 0.55 to 0.70. 
Metric–human agreement improves in parallel and remains close to inter-annotator agreement across all values of $n$. 
This behavior indicates that, on seen data, reducing human noise primarily strengthens agreement signals without revealing large gaps between humans and metrics.

On the unseen PharmaChem domain at $n=1$, the gap between inter-annotator agreement and the best metric–human agreement is small (approximately 0.04), mirroring the segment-level results discussed earlier. 
As we increase $n$, inter-annotator agreement continues to rise and approaches 1.0, reflecting increasingly stable human system rankings. 
In contrast, metric–human agreement plateaus around 0.8. 
As a result, the gap between human–human and human–metric agreement widens, reaching approximately 0.20 for larger group sizes. 
This divergence indicates that once human label inconsistency is reduced, persistent metric weaknesses in unseen domains become more visible.

Grouping also exposes differences between meaningful metrics and trivial baselines. 
At $n=1$, GEMBA performs on par with the sentinel baseline across all domains, indicating that human noise masks the benefit of access to the source sentence. 
When grouping at least $n=6$ segments, this changes: GEMBA outperforms the sentinel by approximately 0.08 pairwise accuracy across all three domains. 
This aligns with the expectation that access to the source sentence improves translation evaluation once annotation noise is reduced. A similar pattern-no separation at $n=1$, but a growing gap after grouping is also observed on the Bio dataset of \citet{zouhar} (see \Cref{apd:acc_bio}).  

For noisy human scores, where strong metrics only marginally outperform a random baseline, we recommend selecting the minimal grouping size $n$ (choosing $n\leq 6$ is often sufficient) that reveals a statistically meaningful separation between a random baseline and a strong metric. By choosing $n$ like this, we remain as close as possible to segment-level evaluation while mitigating human labeling noise.

\subsection{Automatic metrics are susceptible to biases on unseen domains.}
\label{sec:bias1}

In this section, we analyze the systematic biases that occur under domain shifts for various metrics.


\paragraph{Bias 1: QE metrics overestimate open-source systems on unseen domains.}

Any given QE metric should align with human judgments on the relative ranking of MT systems across domains. 
To assess this, we compute system rankings separately for each domain using z-scores. 
These ranks allow us to compare scores across heterogeneous scoring scales of different humans and metrics. For each human and metric, scores are normalized per domain, and systems are ranked by their average score across all translations.

\Cref{fig:human_metric_rankers} shows differences between human- and metric-based rankings for three systems X-ALMA (open-source), Azure, and DeepL (closed-source), for which human judgments reveal performance gaps between open- and closed-source models in en-de.
Human rankings (HumanA/B) are obtained by splitting the four annotators randomly into two groups (WMT23, Emea) or using individual annotators directly (PharmaChem).

On the seen WMT23 domain, humans and metrics largely agree. Both rank DeepL and Azure above X-ALMA (e.g., HumanA/B: DeepL +0.77, Azure +0.62 vs.\ X-ALMA -1.13). Metrics reflect this ordering, with SFT metrics assigning DeepL +1.35, Azure +0.51, and X-ALMA -0.71; Remedy is the only metric placing X-ALMA close to closed-source systems.

On unseen domains, this alignment breaks down. Humans consistently prefer closed-source systems, while metrics inflate scores for X-ALMA and deflate closed-source systems.
On Emea, humans assign DeepL +0.41 and X-ALMA -1.55, whereas SFT metrics score X-ALMA +0.34 and DeepL -1.20, reversing the human ranking.
A similar effect appears on PharmaChem: humans rank DeepL (+0.91) and Azure (+0.40) above X-ALMA (-1.55), while metrics assign X-ALMA scores near zero (e.g., -0.05 for SFT metrics) and push Azure into negative ranges (-1.33).

While the precise score shifts are not all visible in \Cref{fig:human_metric_rankers}, detailed rankings and score differences for all systems are provided in \Cref{apd:system_ranking}.

The bias is most pronounced for supervised fine-tuned metrics, but it also appears for the LLM-as-a-judge approach on the contamination-free PharmaChem domain. 
Notably, these ranking errors occur despite using the same scoring procedures as those used in WMT23, and despite strong inter-annotator agreement among humans.

However, we could not observe the bias for en-zh and en-ko. This effect may be due to several factors. First, shared training data or reward alignment between QE metrics and open-source systems (e.g., X-ALMA) may be stronger for en-de, where more human-annotated data is available, potentially leading to increased system preferences. Second, systems may be more closely matched in quality for en-de, making rankings more sensitive to small scoring differences. Thus, system-specific effects (e.g., relatively weaker performance of X-ALMA in other language pairs) may reduce the visibility of the bias outside en-de.

\begin{figure}[t]
  \includegraphics[width=\columnwidth]{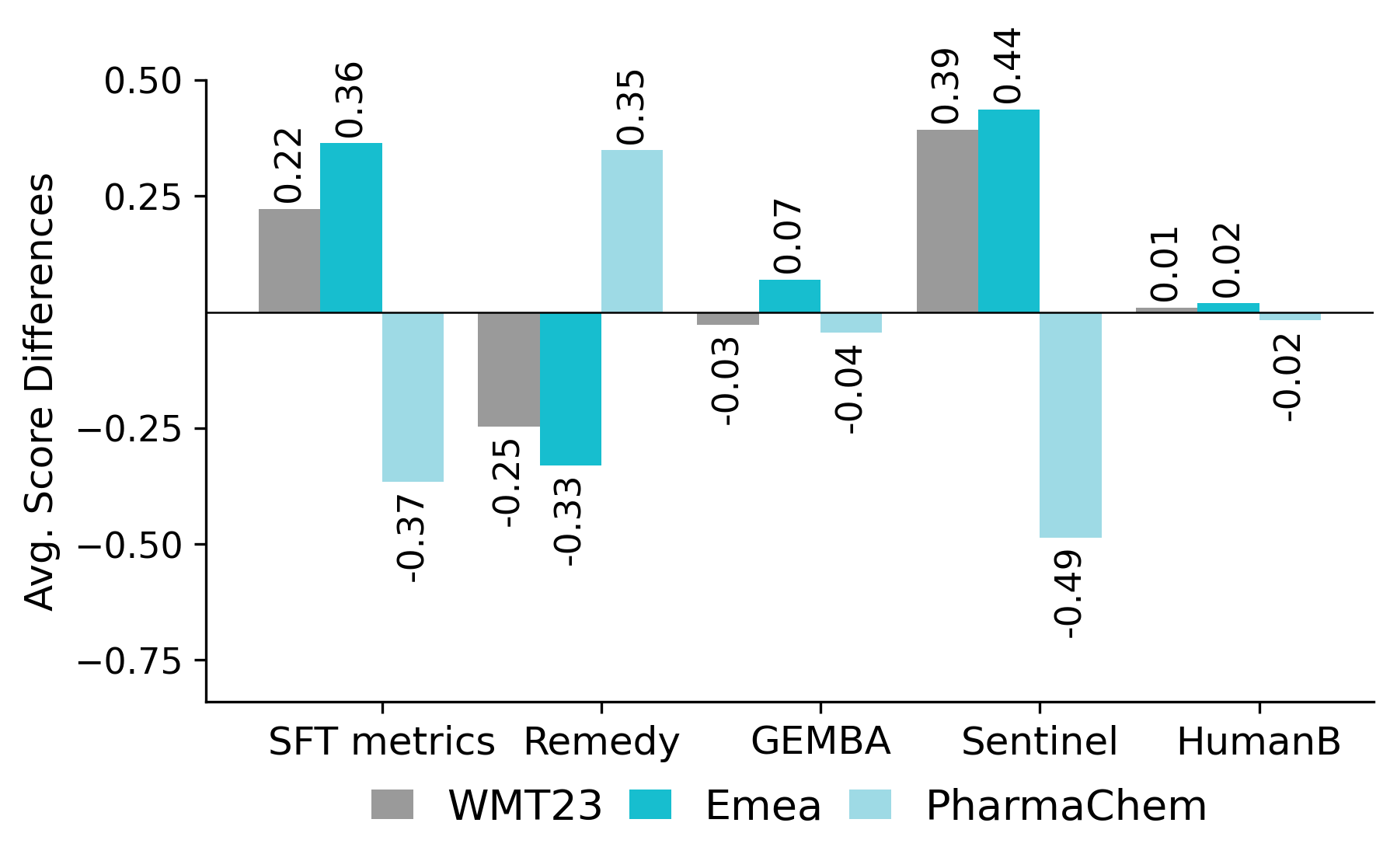}
  \caption{Average differences of normalized human and metric scores across domains. Negative values mean the metric underestimates translation quality relative to humans; positive values mean it overestimates quality.}
  \label{fig:bias2}
  \vspace*{-4mm}
\end{figure}

\paragraph{Bias 2: QE metrics underestimate translation quality on out-of-domain data.}

A desirable property of a QE metric is domain invariance: translation quality should be assessed independently of the source domain.
We test this by comparing average human and metric scores across domains. We use z-scores to ensure comparability across different scoring scales.

For each domain, we compute the average z-score assigned by humans and metrics and  their differences (\Cref{fig:bias2}) over all three language pairs. Because human judgments reflect translation quality rather than domain characteristics, a domain-agnostic metric should produce average scores that closely align with human average scores across domains, resulting in differences near zero.
The results for each pair of languages are in \cref{apd:pessimism_bias}.

\Cref{fig:bias2} shows that on the unseen PharmaChem domain, SFT metrics are systematically pessimistic, assigning scores on average 0.37 points lower than humans. Their behavior closely mirrors that of the sentinel baseline, suggesting reliance on target-side fluency rather than adequacy under domain shift.
Remedy exhibits inconsistent behavior across unseen domains, overestimating quality on PharmaChem while underestimating it on Emea.
In contrast, the LLM-as-a-judge approach (GEMBA) does not show systematic over- or underestimation, indicating more domain-agnostic scoring behavior.

Based on these findings, we recommend using LLM-as-a-judge  when comparing translation quality across domains, as they provide more reliable estimates under domain shift.

\paragraph{Error masking analysis.} 
To better understand why supervised fine-tuned (SFT) metrics exhibit the biases observed on unseen domains, we analyze token-level error masks predicted by \textsc{xComet\_QE}, an explainable variant of \textsc{Comet} \citep{rei2020comet}. Error span prediction offers a fine-grained view of metric behavior beyond scalar scores on the WMT25 Metrics shared task~2.


We compare character-level overlap between \textsc{xComet\_QE}-predicted error spans and human annotations across all domains, computing precision, recall, and F1 following the WMT25 setup. For simplicity, we treat minor and major errors uniformly. Human–human agreement shows balanced precision and recall, with stable F1 scores between 0.37 and 0.47, indicating consistent annotation behavior across domains (\Cref{error_masking_table}).

In contrast, human–\textsc{xComet\_QE} agreement is substantially lower (F1: 0.28–0.33), confirming that error span prediction remains challenging. Precision–recall imbalance is markedly stronger on unseen domains, most notably on PharmaChem, where precision drops to 0.23 while recall reaches 0.80. The imbalance indicates that \textsc{xComet\_QE} over-predicts errors particularly under domain shift. Thus, SFT metrics may rely on source-side domain cues rather than solely judging translation quality. 

Qualitative examples corroborate the pattern: on unseen-domain translations, \textsc{xComet\_QE} frequently labels large portions of a sentence as containing minor errors, even when humans annotate a small number of error spans (\Cref{fig:error_masking_text}). Our analysis reveals a systematic tendency of SFT metrics to over-predict errors on unseen domains, providing an explanation for the pessimistic quality estimates observed in \Cref{fig:bias2}.


\begin{table}[t]
  \centering
    \small
    \begin{tabular}{l@{\hspace{4pt}}cccccc@{\hspace{2pt}}}
      \toprule
      \multirow{2}{*}{\textbf{Domain}} 
        & \multicolumn{3}{c}{\textbf{\textsc{xComet}--human}} 
        & \multicolumn{3}{c}{\textbf{human--human}} \\
      \cmidrule(lr){2-4} \cmidrule(lr){5-7}
        & \textbf{P} & \textbf{R} & \textbf{F1}
        & \textbf{P} & \textbf{R} & \textbf{F1} \\
      \midrule
      WMT23 & 0.27 & \textbf{0.46} & 0.28 & \textbf{0.42} & 0.43 & \textbf{0.37} \\
      Emea           & 0.33 & \textbf{0.51} & 0.33 & \textbf{0.48} & 0.48 & \textbf{0.43} \\
      PharmaChem & 0.23 & \textbf{0.80} & 0.32 & \textbf{0.53} & 0.53 & \textbf{0.47} \\
      \bottomrule
    \end{tabular}
    \caption{\textbf{P}recision, \textbf{R}ecall, and \textbf{F1} of character-level error mask overlap between human--human and human--\textsc{xComet}. \textbf{Bold} shows the highest value per domain.}
    \label{error_masking_table}
  \end{table}

  \begin{figure}[t]
    \centering
   \begin{codeblock}
\textbf{Source:}
Grease line is interrupted near the front end.
\textbf{xCOMET:}
\err{Fettleitung} ist \err{nahe am Front Ende unterbrochen}
\textbf{Human A:}
Fettleitung ist nahe am \err{Front} Ende unterbrochen
\textbf{Human B:}
Fettleitung ist nahe am Front Ende unterbrochen
\end{codeblock}
\vspace{-1em}
\caption{Example of \textsc{xComet} on Emea marking many words as errors when humans only marked two.}
\label{fig:error_masking_text}
\vspace*{-4mm}
 \end{figure}




    


\paragraph{Embedding Analysis.} We analyze last-layer representations of \textsc{xComet}. The embeddings cluster mainly by domain rather than by human-judged quality, indicating reliance on source-side domain information. Details are in \Cref{apd:embeddings PharmaChem}.


\section{Conclusion}




This paper analyzes how automatic MT evaluation metrics behave under domain shift. Using CD-ESA, a new multi-annotator dataset covering translations from the same five MT systems on a seen news domain (WMT23) and two unseen technical domains (Emea and the proprietary PharmaChem), all annotated by the same five annotators, we study metric behavior under controlled cross-domain conditions.

We show that annotation noise can mask metric weaknesses on unseen domains, leading to overly optimistic conclusions. When we reduce noise by averaging multiple human annotations or grouping segments, marked gaps emerge between human and metric judgments on unseen domains.

On unseen domains, metrics show two systematic biases, they underestimating translation quality and over-preferencing weaker open-source systems. These effects are strongest for SFT metrics, while LLM-as-a-judge approaches show greater robustness. Error mask analyses further shows that SFT metrics rely on domain-related cues rather than translation quality.

Our findings highlight that metric robustness across domains must be interpreted relative to inter-annotator agreement under identical conditions. Controlling for human label inconsistency is essential for reliable conclusions about metrics.

\section*{Limitations}

Our study focuses on two unseen technical domains (Emea and PharmaChem) and a single high-resource language pair (English–German), reflecting both our research goal of evaluating metrics in our-of domains settings and the needs of our pharmaceutical industry partner. While extending this approach to additional domains and language pairs is an important direction for future work, the multi-annotator error span annotations required for reliable evaluation are costly at scale. Future studies could mitigate this cost by reusing existing annotations, sharing cross-domain annotation resources, or adopting aggregation strategies such as segment grouping to reduce annotation overhead.

Some of our analyses further rely on assumptions about human judgments. In particular, we assume that human annotators assess translations based on translation quality rather than domain characteristics. Under this assumption, a domain-agnostic metric should assign domain-wise average scores similar to those of humans. To our knowledge, this assumption is necessary for fair cross-domain comparison, but our conclusions may need to be revisited if future work challenges it.


Finally, reducing human annotation noise requires multiple annotations per segment, which substantially increases annotation cost. While averaging human judgments improves evaluation reliability, it is expensive in practice. Our proposed grouping strategy offers a more cost-effective alternative, but it reduces the number of independent evaluation units. To mitigate this trade-off, future evaluation campaigns could prioritize annotating shorter segments, which would allow collecting more annotations under fixed annotation budgets.

\section*{Acknowledgments}

This work was supported by the Lower Saxony Ministry of Science and Culture and the VW Foundation, and by
the Federal Ministry for Economic Affairs and Climate Action (BMWK) on the basis of a decision by the German Bundestag.

The authors would like to thank the German Federal Ministry of Research, Technology and
Space and the German federal states (\url{http://www.nhr-verein.de/en/our-partners}) for
supporting this work/project as part of the National High-Performance Computing (NHR) joint
funding program.

We thank Anastasia Zhukova for supporting on the data curation and preparation phase and eschbach GmbH, specifically Christian E. Lobmüller, for funding the data annotation effort to evaluate the approach on the proprietary domain.

We thank Tianyu Yang for carefully reading the final draft of the paper and providing valuable feedback. 
Finally, we thank Lars Kaesberg and Jonas Becker for their support in improving the figure design. 


\bibliography{custom,anthology}

\appendix



\section{Pairwise accuracy evaluation for the BIO-MQM dataset}
\label{apd:acc_bio}

Prior work by \citet{zouhar} evaluates metric robustness on the BIO-MQM dataset using plain Kendall’s~$\tau$ for meta-evaluation. While Kendall’s~$\tau$ is a widely used rank-correlation measure, it does not explicitly account for ties, which are common in high-quality translation data.

More recent meta-evaluation techniques address these issues by (i) comparing systems on a pairwise basis with explicit tie handling \citep{deutsch2023ties}, and (ii) restricting comparisons to translations of the same source segment via a group-by-item evaluation strategy. Together, these methods prevent measuring spurious correlations across heterogeneous segments and consider if metrics are able to predict ties.

Applying these updated evaluation techniques to the BIO-MQM data yields a more nuanced view of metric performance and highlights several properties of the dataset that influence the resulting conclusions.

\begin{figure*}[t]
  \includegraphics[width=\textwidth]{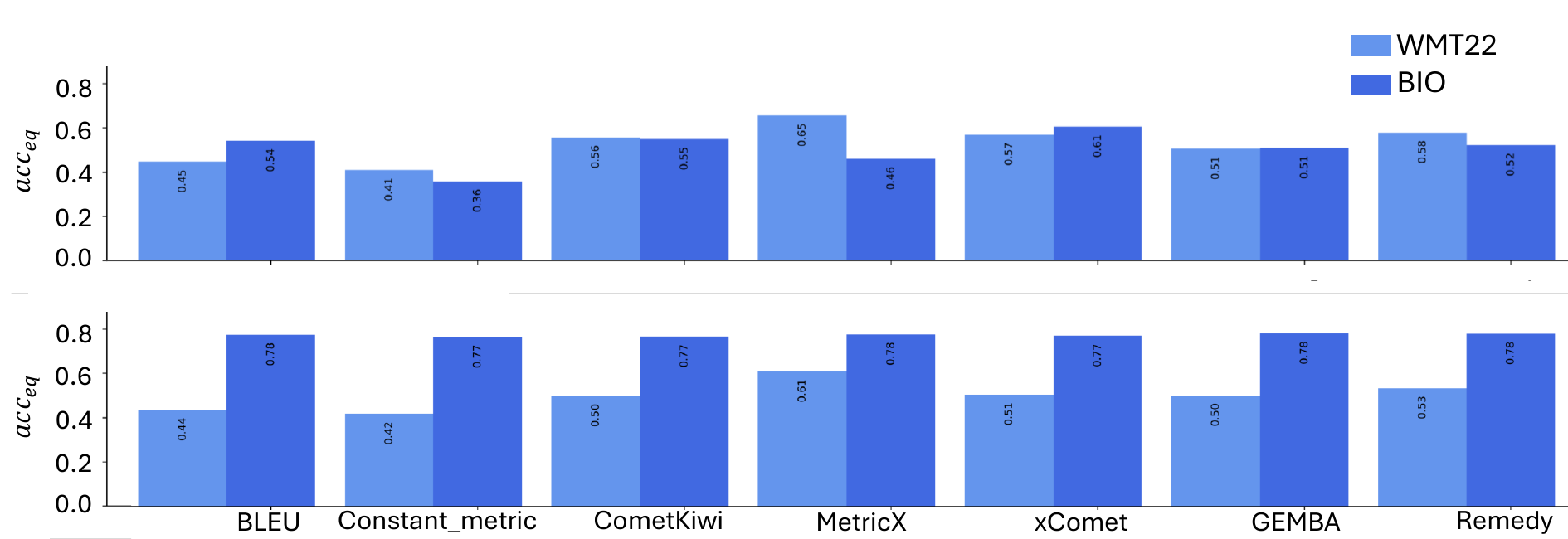}
  \caption{Pairwise metric--human accuracy on BIO-MQM data from WMT22. Except for BLEU, which requires references, all metrics are evaluated in a reference-free (QE) setting. The top panel shows results for English--Russian and the bottom panel for Chinese--English.}
  \label{fig:acc_plain_bio}
\end{figure*}

\paragraph{Findings from the updated BIO meta-evaluation.}
Figure~\ref{fig:acc_plain_bio} summarizes metric-human agreement under pairwise accuracy with tie calibration.

First, for the Chinese--English BIO data, performance is strongly affected by dataset characteristics. A large proportion of translations contain no annotated errors, leading to label imbalance. Under these conditions, no metric consistently outperforms a constant baseline that predicts ties for all system pairs. This suggests that apparent metric failures on this subset are driven primarily by data properties rather than by a lack of metric signal.

Second, BLEU performs unexpectedly well on the BIO domain, particularly for English--Russian, where it exceeds the constant baseline by nearly 20 accuracy points. On WMT22 news data, in contrast, BLEU performs only marginally above this baseline. This pattern indicates that BLEU’s behavior is domain-dependent: it appears to benefit from the more literal and formulaic nature of biomedical text, whereas it is less effective on the more diverse and conversational content found in WMT news data. As a result, BLEU cannot be assumed to act as a domain-invariant reference metric.

Finally, we note a structural limitation of the BIO-MQM dataset as used in prior work. Only a small number of segments are translated by the same systems and annotated by at least two human annotators, which makes it difficult to compute reliable inter-annotator agreement or human-oracle upper bounds. This limits the interpretability of absolute metric performance levels on this dataset.

\paragraph{Grouping analysis on BIO data.}

\begin{figure}[t]
  \includegraphics[width=\columnwidth]{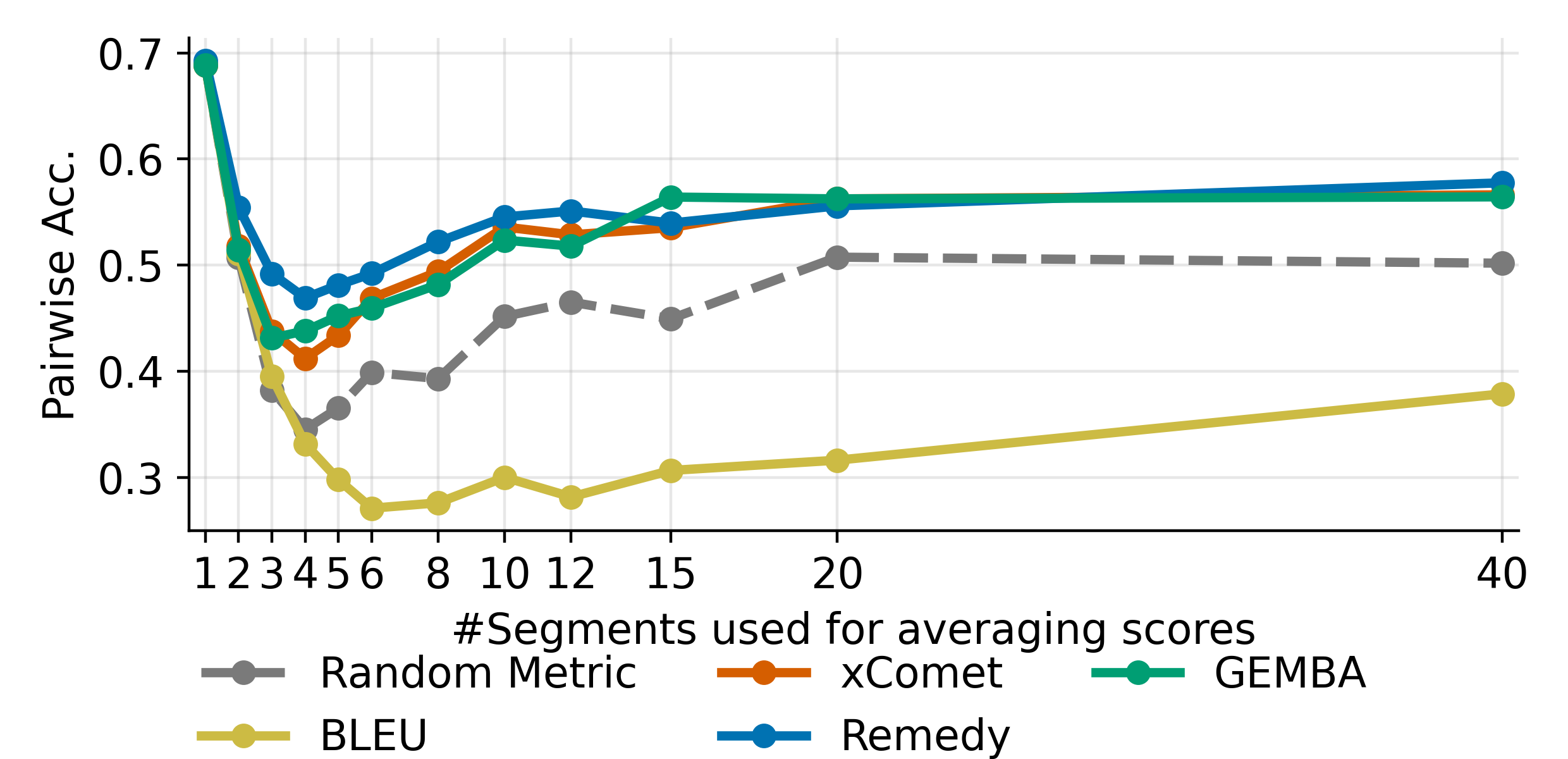}
  \caption{Pairwise accuracy on Chinese--English BIO data when grouping multiple segments into a single evaluation unit.}
  \label{fig:averaging_bio}
\end{figure}

To further assess whether label imbalance and annotation noise drive these effects, we apply our proposed grouping $n$ segments strategy described in \Cref{sec:averaging}. As shown in Figure~\ref{fig:averaging_bio}, grouping a small number of segments substantially stabilizes metric evaluation on the Chinese--English BIO data without discarding any annotations.

Under grouping, most metrics outperform a random baseline, indicating that earlier near-random performance was largely due to noise and imbalance at the segment level rather than an absence of useful metric signal. BLEU, however, continues to lag behind neural metrics under grouping, consistent with its known limitations in fine-grained quality estimation.

Overall, this analysis illustrates that conclusions about metric robustness on unseen domains are sensitive to the choice of meta-evaluation protocol and to dataset properties such as label balance and annotation density. These observations motivate the controlled evaluation design adopted in our main experiments.

\section{Meta-evaluation details}
\label{apd:meta-eval}
\paragraph{Segment-level.} For a given segment, let $C$ denote the number of translation pairs ranked in the same order by both $m$ and $h$, and $D$ the number ranked in opposite order. Let $T_m$ and $T_h$ denote pairs tied only by the metric or only by the human, and $T_{mh}$ pairs tied by both. Segment-level accuracy is then defined as:
\begin{equation}
\mathrm{acc}_{eq} =
\frac{C + T_{mh}}{C + D + T_m + T_h + T_{mh}}.
\end{equation}
Tie calibration introduces an optimal threshold on metric score differences (i.e., the maximum score difference still treated as a tie), so that metrics producing continuous scores and rarely predicting exact ties can still be fairly compared to human judgments.

\paragraph{System-level.} Given a test set with \(M\) MT systems, SPA considers all \(\binom{M}{2}\) system pairs. For each pair \((i,j)\), we compute a human p-value \(p^{h}_{ij}\) and a metric p-value \(p^{m}_{ij}\), representing the statistical confidence that system \(i\) is preferred over system \(j\). These p-values are obtained via paired permutation tests over segment-level scores. SPA assigns partial credit based on how similar the metric’s confidence is to the human confidence:
\begin{equation}
\mathrm{SPA} = \binom{M}{2}^{-1}
\sum_{i=0}^{M-1} \sum_{j=i+1}^{M-1}
\left( 1 - \left| p^{h}_{ij} - p^{m}_{ij} \right| \right).
\end{equation}
Thus, SPA rewards metrics that express preference strengths similar to those of human judgments and penalizes deviations.

\section{Human annotation times}
\label{apd:human_annotation_stats}

\begin{figure}[t]
  \includegraphics[width=\columnwidth]{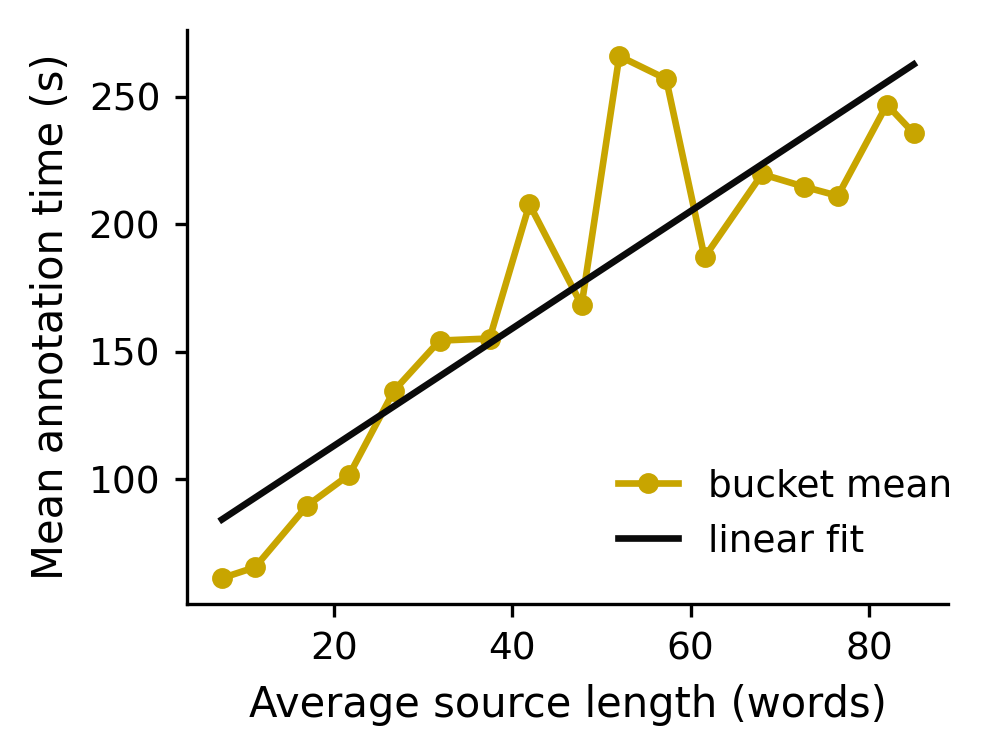}
  \caption{Human annotation time as a function of source segment length. Source sentences are bucketed by length in intervals of five words. The average segment length (x-axis) and annotation time (y-axis) is reported per bucket. The mean is computed over the 75\% fastest annotations within each bucket.}

  \label{fig:annotation_time}
\end{figure}

Figure~\ref{fig:annotation_time} analyzes how human annotation time scales with source segment length. We bucket source sentences by length (in word intervals of five) and report the average annotation time per bucket. To reduce the influence of outliers (e.g., breaks during annotation), we compute the mean over the 75\% fastest annotations within each bucket.

The results show that annotation time increases approximately linearly with segment length. These finding supports the recommendation made in the main text: annotating shorter segments is more cost-efficient. When combined with grouping strategies that aggregate multiple short segments into a single evaluation unit, this approach enhances evaluation robustness while limiting annotation cost.

\begin{figure}[h]
  \centering

  \begin{subfigure}{\columnwidth}
    \centering
    \includegraphics[width=\linewidth]{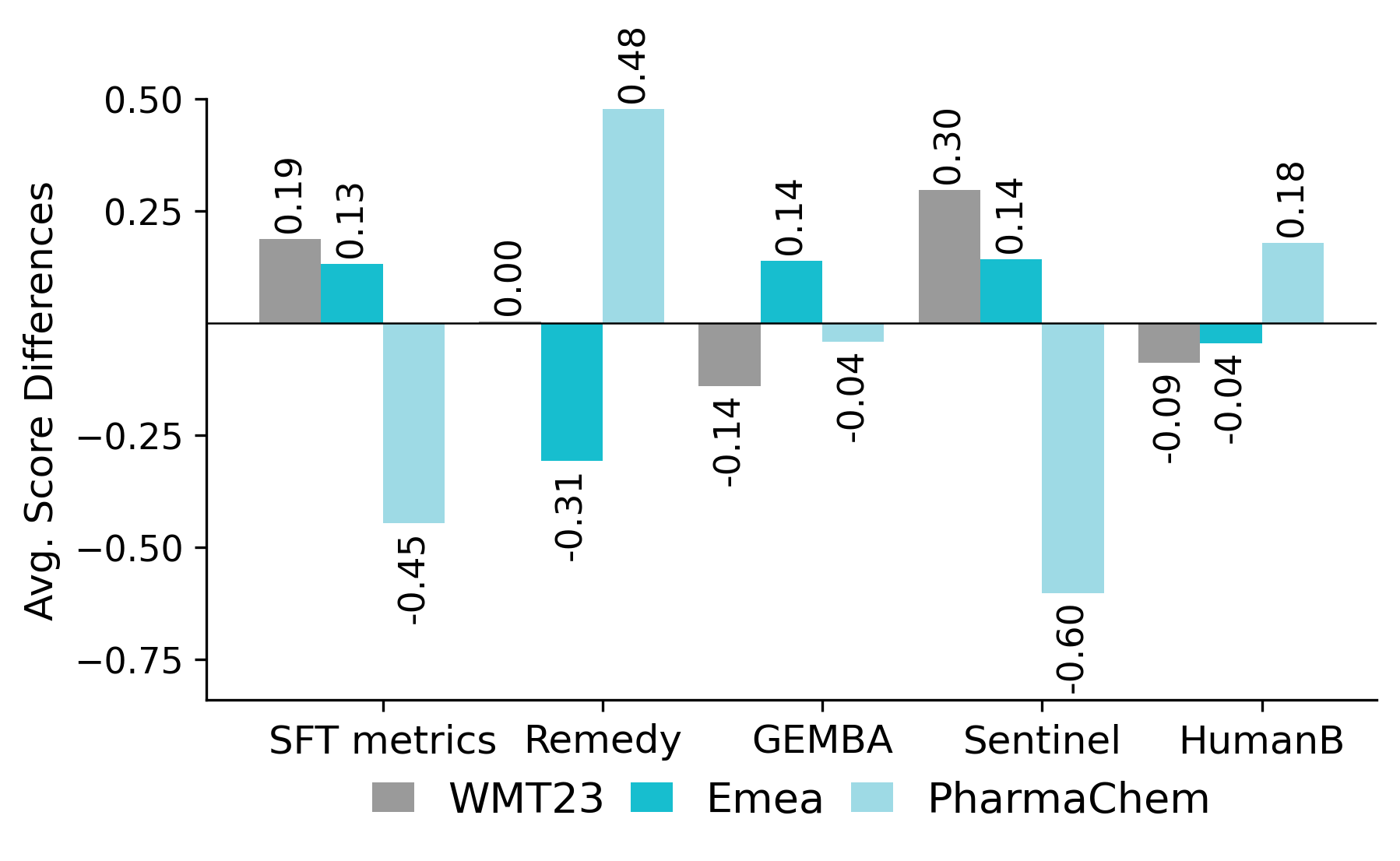}
    \caption{English-German}
  \end{subfigure}

  \begin{subfigure}{\columnwidth}
    \centering
    \includegraphics[width=\linewidth]{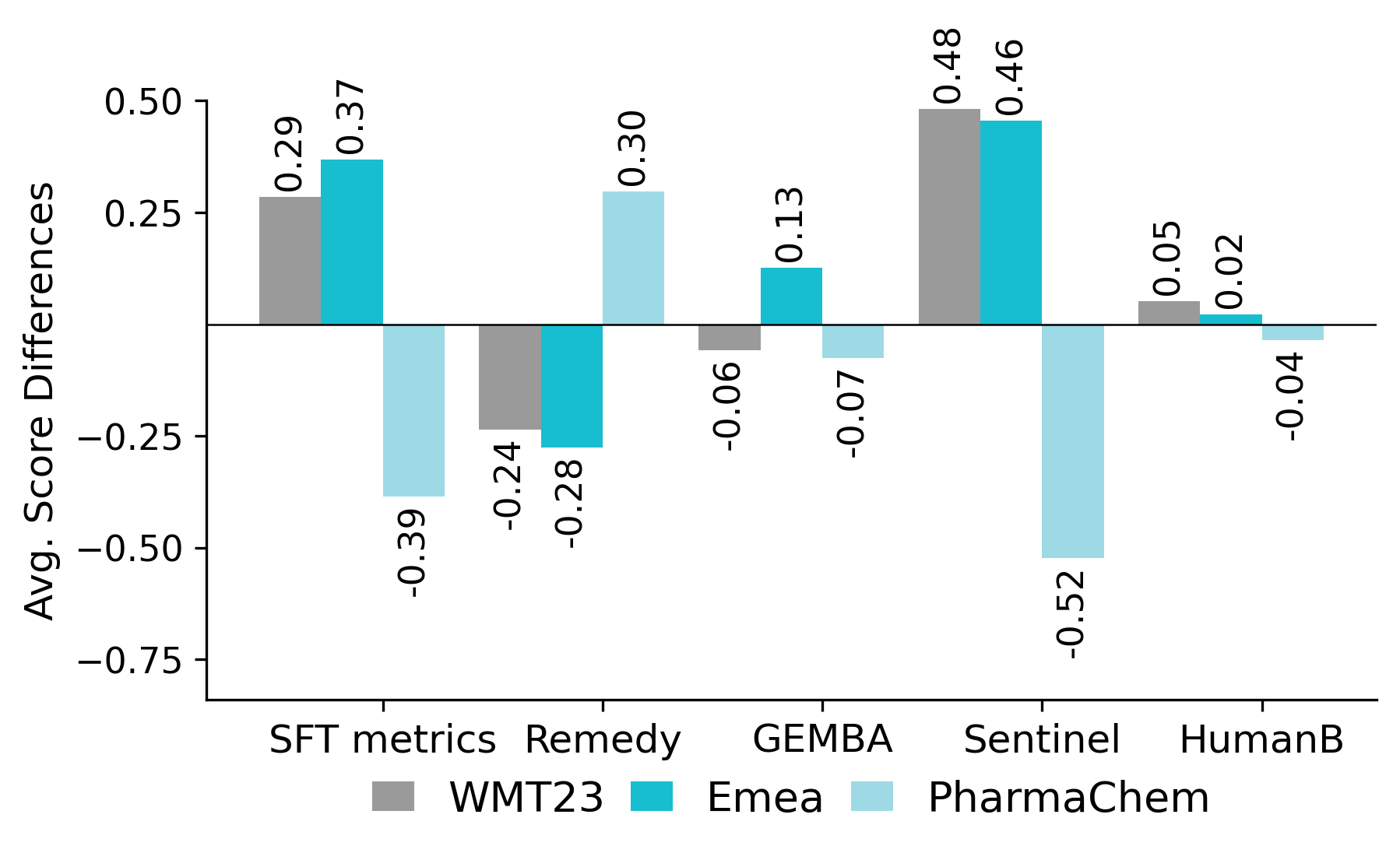}
    \caption{English-Korean}
  \end{subfigure}

  \begin{subfigure}{\columnwidth}
    \centering
    \includegraphics[width=\linewidth]{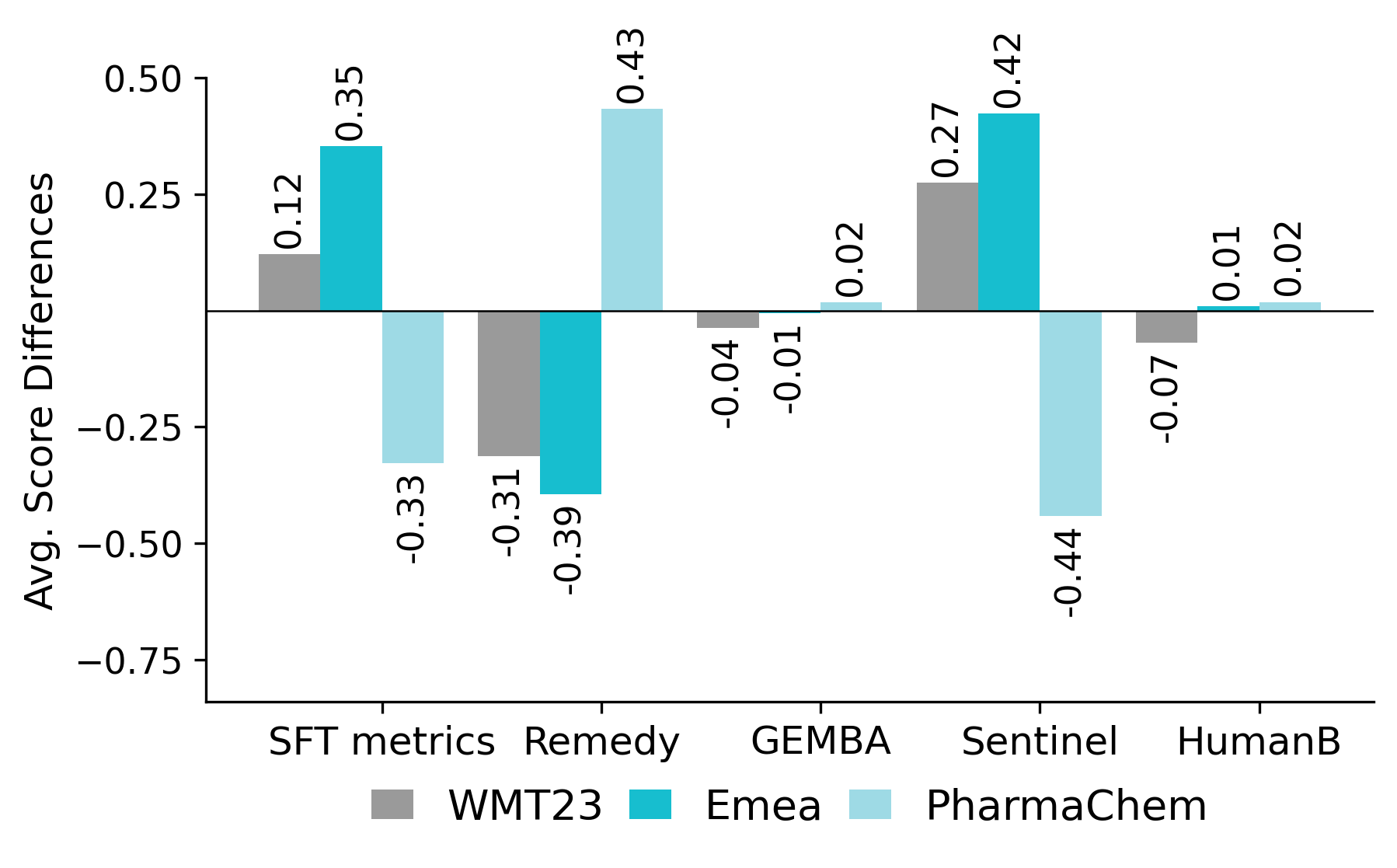}
    \caption{English-Chinese}
  \end{subfigure}

  \caption{Average differences of normalized human and metric scores across domains. Negative values mean the metric underestimates translation quality relative to humans; positive values mean it overestimates quality.}
  \label{fig:bias2_3lps}
  \vspace*{-4mm}
\end{figure}

\section{Over pessimism bias of QE metrics on unseen domains across three langauge pairs }
\label{apd:pessimism_bias}

Following \Cref{fig:bias2}, we compute, for each domain, the difference between average human and metric z-scores to assess whether metrics systematically over- or underestimate translation quality. Instead of averaging over all three language pairs, we report results for each language pair in \Cref{fig:bias2_3lps}. Overall, the patterns observed in the main text are consistent across all language pairs. 

On the unseen PharmaChem domain, SFT metrics consistently underestimate translation quality relative to human judgments. For all three langauge pairs the SFT metrics score the translation on the PharmaChem between 0.3 to 0.45 below human scores. Notably, the behavior of SFT metrics closely follows that of the \texttt{sentinel\_cand} baseline across all three language pairs.

Remedy again shows unstable behavior across domains. For instance, it overestimates quality on PharmaChem across all three language pairs by 0.30 to 0.48 points and underestimates translation quality by 0.28-0.39 on Emea. In contrast, GEMBA exhibits more stable behavior, with differences to human scores remaining closer to zero across domains and language pairs (typically within $\pm$0.01–0.15), indicating more domain-agnostic scoring.

These consistent trends across English–German, English–Chinese, and English–Korean suggest that the observed pessimism bias of supervised QE metrics is not language-specific but a general phenomenon under domain shift.

\section{Embeddings of \textsc{xComet} }
\label{apd:embeddings PharmaChem}

To understand why SFT metrics become overly pessimistic on unseen domains, we analyze what information their internal representations encode. 
Specifically, we extract the last-layer embeddings of \textsc{xComet} before the scoring head and visualize translation embeddings from WMT23 and PharmaChem using t-SNE.

Ideally, these embeddings would primarily reflect translation quality.
Good and bad translations should form separate clusters, and these groupings should be largely independent of the domain of the source sentence.

\begin{figure}[t]
  \centering

  \begin{subfigure}{\columnwidth}
    \centering
    \includegraphics[width=\linewidth]{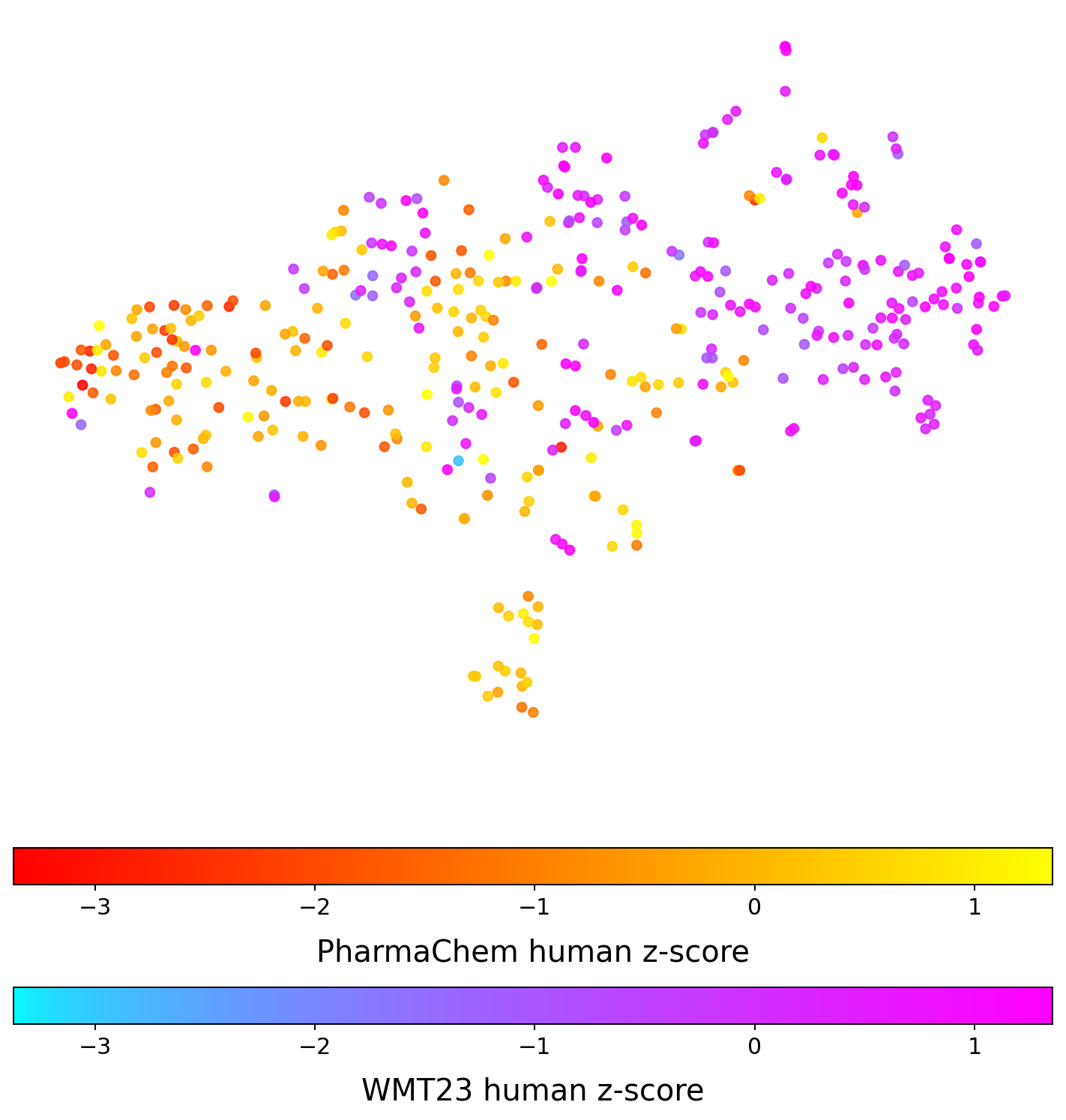}
    \caption{PharmaChem (red--yellow) vs.\ WMT (blue--violet).}
    \label{fig:embeddings_sc_shift}
  \end{subfigure}

  \vspace{0.6em}

  \begin{subfigure}{\columnwidth}
    \centering
    \includegraphics[width=\linewidth]{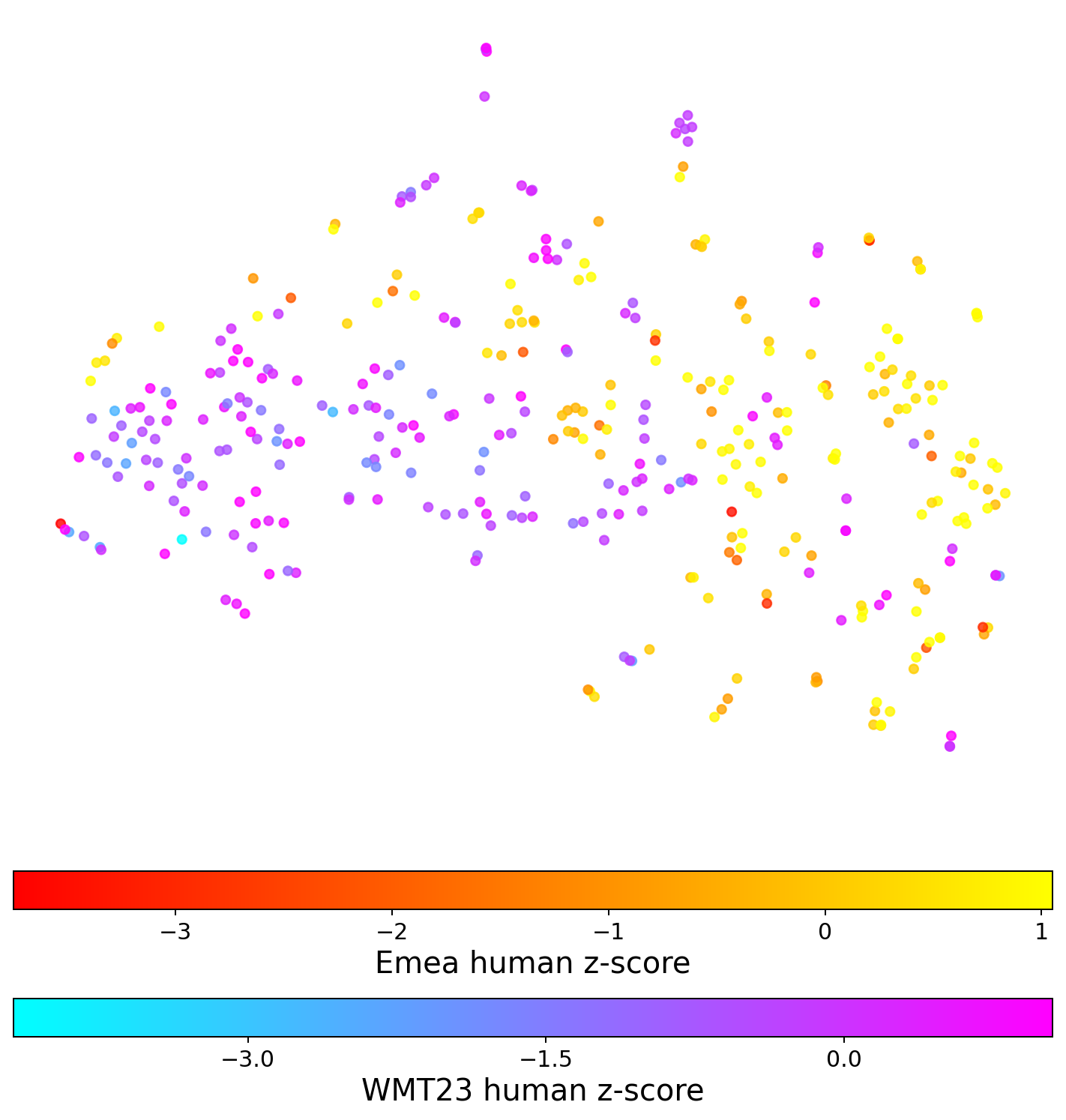}
    \caption{Emea (red--yellow) vs.\ WMT (blue--violet).}
    \label{fig:embeddings_sc_emea}
  \end{subfigure}

  \caption{t-SNE visualization of last-layer translation embeddings from \textsc{xComet} (before the linear scoring head). Each point represents a translation, colored by human-assigned z-scores. }
  \label{fig:embeddings_sc}
\end{figure}

\Cref{fig:embeddings_sc} shows the resulting embedding space, where each point corresponds to a translation, and colormaps indicate the source domain. 
Rather than clustering by human-judged translation quality, the embeddings separate almost entirely by domain. 
Translations from WMT23 and PharmaChem exhibit a clear domain-dependent tendency, clustering more by domain than by human-judged quality. 
This pattern indicates that domain information is unsurprisingly still encoded, which could partially explain reduced robustness under domain shift.

These results indicate that \textsc{xComet} learns strong domain-specific representations. As a consequence, the metric relies heavily on source-side domain information, which weakens its ability to generalize and to judge translation quality consistently across domains.

We observe the same behavior when repeating the analysis for WMT23 and Emea (details in \Cref{apd:embeddings PharmaChem}). 
In all cases, domain-specific clustering persists, largely independent of translation quality.

\section{Data filtering}
\label{apd:Domain_Data_filter}
To obtain a close proxy to our private company dataset, we manually searched for domain-related multilingual and parallel open-source resources and trained a domain classifier to filter and retain only those segments that closely match the chemical and pharmaceutical domain.
To train a domain classifier for identifying chemical and pharmaceutical content, we constructed a dataset with a balanced mixture of related and unrelated segments. Half of the data (50\%) consists of unrelated domains such as Law, Art, Physics, and History, sampled from the EU Bookshop corpus.\footnote{\url{https://metatext.io/datasets/eubookshop}} The other half (50\%) contains domain-related text. Of this portion, 50\% is drawn from FoS categories Medicine, Pharma \citep{singh2023scirepeval}, and 50\% consists of English instances from the PharmaChem domain, sampled randomly to ensure representation across all customers.

The final dataset contains 80K documents, split into train, dev, and test sets. We fine-tuned SciBERT \citep{beltagy2019scibert} using the following hyperparameters:

\begin{verbatim}
- learning_rate=2e-5,
- batch_size=16,
- num_train_epochs=3,
- weight_decay=0.01,
- evaluation_strategy=epoch,
- save_strategy=epoch,
- load_best_model_at_end=True
\end{verbatim}

We observed overfitting beyond 3 epochs and therefore kept the best checkpoint. The final model achieved accuracy~0.95, precision~0.97, recall~0.93, and F1~0.95.

We qualitatively inspected several multilingual datasets for domain relevance, including XQuAD \citep{xquad} (Wikipedia QA pairs), TED Talks \citep{ted} (semi-formal spoken-language transcripts), and Emea \citep{emea} (medical documents from the European Medicines Agency). Among these, we found that Emea is the closest in style and domain to the PharmaChem domain. After training, we then applied the classifier to the Emea corpus and retained only sentences with a classifier score $> 0.5$ to obtain a proxy dataset for the private company dataset.

\section{System ranking table }
\label{apd:system_ranking}

This appendix reports the full system rankings for english-german underlying the analyses in \Cref{sec:bias1}.  
For each domain, we show rankings assigned by individual human groups (HumanA, HumanB) and by each evaluated metric. Rankings are computed from domain-wise z-normalized scores, with higher values indicating better performance.

The tables make explicit where metrics diverge from human judgments, including cases of attenuated score differences and outright ranking reversals between open- and closed-source systems. They also provide the complete score context for systems not visualized in the main figures, enabling a detailed inspection of metric behavior across domains.

\begin{table*}[t]
  \centering
  \small

  \begin{subtable}{\linewidth}
    \centering
    \begin{tabular}{ccccc}
      \toprule
      \textbf{humanA} & \textbf{humanB} & \textbf{SFT metrics} & \textbf{GEMBA} & \textbf{Remedy} \\
      \midrule
       GPT4o (+0.94) &
      GPT4o (+1.00) &
      DeepL (+1.35) &
      GPT4o (+1.11) &
      Azure (+0.62) \\

      DeepL (+0.77) &
      DeepL (+0.94) &
      Azure (+0.51) &
      Azure (+0.89) &
      DeepL (+0.60) \\

      Azure (+0.62) &
      Azure (+0.31) &
      GPT4o (+0.32) &
      DeepL (+0.11) &
      \cellcolor{orange!25} GPT4o (+0.44) \\
        \midrule
      ALMA (-1.13) &
      Tower-Instrc (-0.86) &
      ALMA (-0.71) &
      Tower-Instrc (-0.45) &
      \cellcolor{orange!25} ALMA (+0.32) \\

      Tower-Instrc (-1.19) &
      ALMA (-1.40) &
      Tower-Instrc (-1.46) &
      ALMA (-1.66) &
      Tower-Instrc (-1.99) \\
      \bottomrule
    \end{tabular}
    \caption{WMT23 (en--de).}
  \end{subtable}

  \vspace{0.8em}

  \begin{subtable}{\linewidth}
    \centering
    \begin{tabular}{ccccc}
      \toprule
      \textbf{humanA} & \textbf{humanB} & \textbf{SFT metrics} & \textbf{GEMBA} & \textbf{Remedy} \\
      \midrule
      Azure (+0.76) &
      Azure (+1.41) &
      GPT4o (+1.32) &
      DeepL (+0.98) &
      Azure (+1.14) \\

      GPT4o (+0.74) &
      \cellcolor{green!20} DeepL (+0.41) &
      Azure (+0.48) &
      Azure (+0.83) &
      \cellcolor{red!25}ALMA (+0.59) \\

      \cellcolor{green!20} DeepL (+0.41) &
      GPT4o (+0.26) &
      \cellcolor{red!25} ALMA (+0.34) &
      GPT4o (+0.56) &
      GPT4o (+0.56) \\

      Tower-Instrc (-0.36) &
      Tower-Instrc (-0.53) &
      Tower-Instrc (-0.94) &
      Tower-Instrc (-0.86) &
      \cellcolor{red!25} DeepL (-0.67) \\

      \cellcolor{green!20} ALMA (-1.55) &
      \cellcolor{green!20} ALMA (-1.55) &
      \cellcolor{red!25} DeepL (-1.20) &
      ALMA (-1.51) &
      Tower-Instrc (-1.62) \\
      \bottomrule
    \end{tabular}
    \caption{Emea (en--de).}
  \end{subtable}

  \vspace{0.8em}

  \begin{subtable}{\linewidth}
    \centering
    \begin{tabular}{ccccc}
      \toprule
      \textbf{humanA} & \textbf{humanB} & \textbf{SFT metrics} & \textbf{GEMBA} & \textbf{Remedy} \\
      \midrule
      GPT4o (+1.00) &
      GPT4o (+1.26) &
      GPT4o (+1.48) &
      GPT4o (+1.70) &
      GPT4o (+1.15) \\

      DeepL (+0.91) &
      DeepL (+0.56) &
      DeepL (+0.52) &
      DeepL (+0.32) &
      DeepL (+0.95) \\

      \cellcolor{green!20} Azure (+0.40) &
      \cellcolor{green!20} Azure (+0.55) &
      \cellcolor{red!25} ALMA (-0.05) &
      \cellcolor{red!25} ALMA (-0.09) &
      \cellcolor{red!25} ALMA (-0.05) \\

      Tower-Instrc (-0.77) &
      \cellcolor{green!20} ALMA (-1.09) &
      Tower-Instrc (-0.62) &
      \cellcolor{red!25} Azure (-0.81) &
      \cellcolor{red!25} Azure (-0.44) \\

      \cellcolor{green!20} ALMA (-1.55) &
      Tower-Instrc (-1.27) &
      \cellcolor{red!25} Azure (-1.33) &
      Tower-Instrc (-1.16) &
      Tower-Instrc (-1.61) \\
      \bottomrule
    \end{tabular}
    \caption{PharmaChem (en--de).}
  \end{subtable}

  \caption{Human and metric rankings of MT systems across three domains.
Columns show rankings produced by each human group or metric; values in parentheses are average domain-wise z-normalized scores.
\colorbox{green!20}{Green} highlights system pairs where humans agree but metrics do not.
\colorbox{orange!25}{Orange} indicates cases where metrics preserve the human-preferred ordering between open- and closed-source systems but with substantially smaller score gaps.
\colorbox{red!25}{Red} marks ranking reversals relative to human judgments.
For WMT23 and Emea, four annotators were split into two pairs (HumanA/HumanB); for PharmaChem, two annotators were used directly.
}
  \label{tab:domain_sft_comparison}
\end{table*}

\section{Human annotation guidelines}
Guidelines and Labeling setup were adapted from error span annotation protocol (ESA, \citep{kocmi2024error}).
Provided guidelines: 
\subsection*{Introduction}
We want to find out which translation system on the shift connector data is mostly preferred by humans, and if automatic evaluation metrics come to the same result as human evaluation. For that, translations of English source sentences were done by six different machine translation systems (Deepl, GPT4-o, Azure Translate, TowerInstruct-7B-v0.2, Tower fine-tuned on PharmaChem data, and X-ALMA), which are known to produce high-quality translations. You are asked to annotate the quality of the translated sentences.

\subsection*{Guidelines}
Annotate the English-German translation by marking errors and providing a translation quality score:
\begin{itemize}
    \item Choose error severity for translation:
    \begin{itemize}
        \item Minor errors: Style, grammar, and word choice could be better or more natural.
        \item Major errors: The meaning is changed significantly, and/or the part is really hard to understand.
    \end{itemize}
    \item Missing content: If something is missing, tick either the “Minor Omission” or “Severe Omission” box. If not, tick the “No Omission” box
    \item Tip: Highlight the word or general area of the error---it doesn’t need to be exact. Use separate highlights for different errors.
    \item Score the translation: After marking errors, please set an overall score based on meaning preservation and general quality:
    \begin{itemize}
        \item 0: No meaning preserved: most information is lost.
        \item 33\%: Some meaning preserved: major gaps and narrative issues.
        \item 66\%: Most meaning preserved: minor issues with grammar or consistency.
        \item 100\%: Perfect: meaning and grammar align completely with the source
    \end{itemize}
\end{itemize}
Tip: Assign the score such that it reflects your preferred translations. If you prefer translation of system 1 over the other two translations, the score of translation 1 should be the highest.

\subsection*{Additional Notes}
\begin{itemize}
    \item If you don’t know an abbreviation or word, quickly Google it or ask Google Gemini.
    \item Do not use ChatGPT if you don’t know a word, as we use ChatGPT as one of the translation systems. It probably gives you the same (but maybe wrong) translation.
    \item Some source sentences will appear twice. This is done on purpose for validation reasons. Proceed as always.
    \item The order of the translation systems is randomly shuffled. For example, the translation of system 1 could be the GPT4-o translation for one example and in the next example, system 1 could be Deepl.
\end{itemize}

\begin{figure*}[t]
    \centering
    \includegraphics[width=\textwidth]{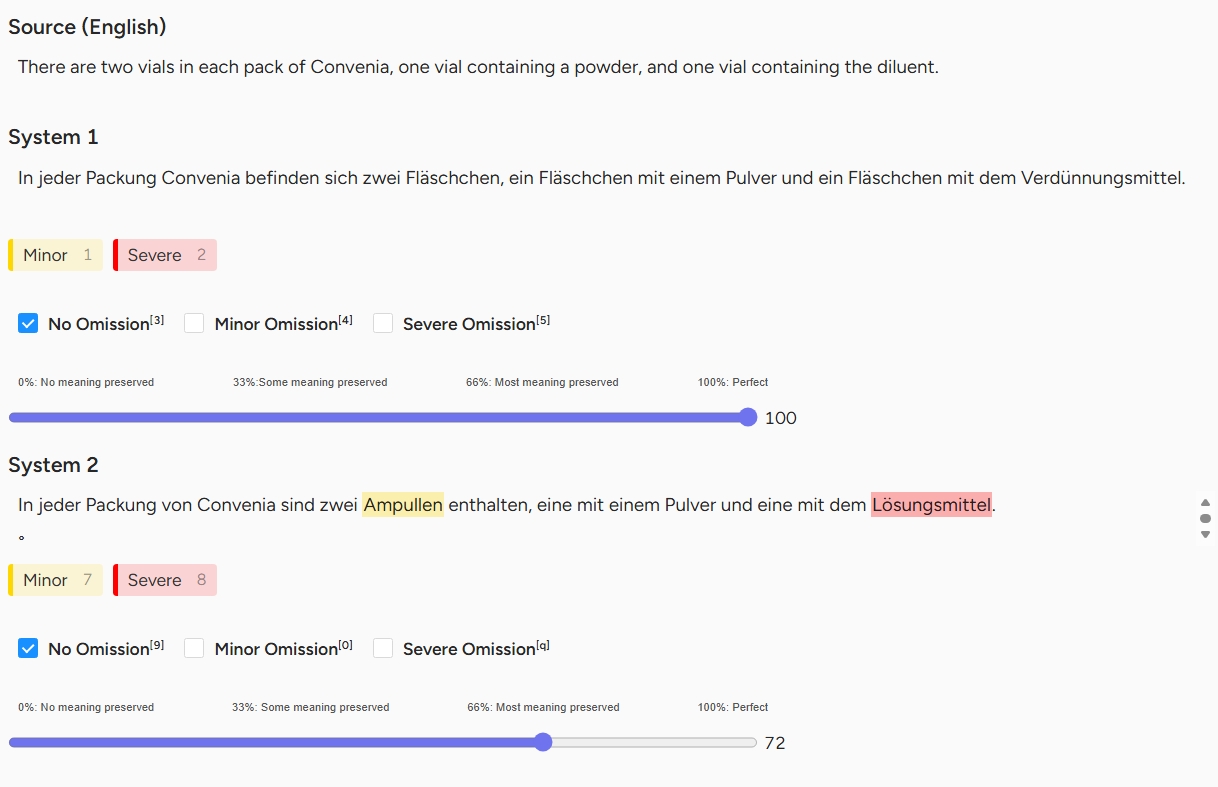}
    \caption{Example for the annotation interface in Label Studio.}
    \label{fig:annotation_interface_img}
\end{figure*}

Annotation interface is adapted from ESA annotation interface \footnote{https://github.com/AppraiseDev/Appraise} and rebuild in Label-Studio. An image of the rebuild interface is given in \Cref{fig:annotation_interface_img}.

\section{AI usage card}

\Cref{tab:aicard} shows how AI was used to construct this study apart from using AI models as subjects of experiments (e.g., for writing, writing code).

\makeAIUsageCard

\end{document}